\documentclass[sigconf]{acmart}
\usepackage[ruled, boxed]{algorithm2e}
\usepackage{color}
\newtheorem{definition}{Definition}

\AtBeginDocument{%
  \providecommand\BibTeX{{%
    \normalfont B\kern-0.5em{\scshape i\kern-0.25em b}\kern-0.8em\TeX}}}

\setcopyright{rightsretained}
\copyrightyear{2023}
\acmYear{2023}
\acmConference[CIKM '23] {Proceedings of the 32nd ACM International Conference on Information and Knowledge Management}{October 21--25, 2023}{Birmingham, United Kingdom.}
\acmBooktitle{Proceedings of the 32nd ACM International Conference on Information and Knowledge Management (CIKM '23), October 21--25, 2023, Birmingham, United Kingdom}
\acmPrice{}
\acmISBN{979-8-4007-0124-5/23/10}
\acmDOI{10.1145/3583780.3615104}

\DeclareUnicodeCharacter{2212}{-}

\usepackage{multirow}
\usepackage{graphicx}
\begin{document}

\title{Unveiling the Role of Message Passing in Dual-Privacy Preservation on GNNs}

\author{Tianyi Zhao}
\affiliation{%
  \institution{University of Southern California}
  \city{Los Angeles}
  \state{California}
  \country{USA}}
\email{azaz010713@gmail.com}

\author{Hui Hu}
\affiliation{%
  \institution{University of Wyoming}
  \city{Laramie}
  \state{Wyoming}
  \country{USA}}
\email{huihu0804@gmail.com}

\author{Lu Cheng}
\affiliation{%
  \institution{University of Illinois Chicago}
  \city{Chicago}
  \state{Illinois}
  \country{USA}}
\email{lucheng@uic.edu}

\begin{abstract}
Graph Neural Networks (GNNs) are powerful tools for learning representations on graphs, such as social networks. However, their vulnerability to privacy inference attacks restricts their practicality, especially in high-stake domains. To address this issue, privacy-preserving GNNs have been proposed, focusing on preserving node and/or link privacy. This work takes a step back and investigates how GNNs contribute to privacy leakage. Through theoretical analysis and simulations, we identify message passing under structural bias as the core component that allows GNNs to \textit{propagate} and \textit{amplify} privacy leakage. Building upon these findings, we propose a principled privacy-preserving GNN framework that effectively safeguards both node and link privacy, referred to as dual-privacy preservation. The framework comprises three major modules: a Sensitive Information Obfuscation Module that removes sensitive information from node embeddings, a Dynamic Structure Debiasing Module that dynamically corrects the structural bias, and an Adversarial Learning Module that optimizes the privacy-utility trade-off. Experimental results on four benchmark datasets validate the effectiveness of the proposed model in protecting both node and link privacy while preserving high utility for downstream tasks, such as node classification.

\end{abstract}



\maketitle

\section{Introduction}
In the past few years, Graph Neural Networks (GNNs) have experienced significant advancements and have found widespread application across various domains, including critical areas like financial analysis \cite{cheng2022financial,yang2021financial}, traffic predictions \cite{wang2020traffic,diehl2019graph,wang2023causalse}, and drug discovery \cite{gaudelet2021utilizing}. Despite its superior performance in various learning tasks, GNN is found to be extremely vulnerable to privacy inference attacks, such as node attribute inference attacks \cite{zhang2022inference,duddu2020quantifying}, link inference attacks \cite{wu2022linkteller}, and membership inference attacks \cite{olatunji2021membership,he2021node,wu2021adapting}. 
In response, a surge of privacy-preserving GNNs have been proposed to protect users' privacy w.r.t., for example, nodes' sensitive attributes \cite{zhou2020privacy,hu2022learning} and/or the link status \cite{wu2022linkteller,he2020stealing}. Therefore, what are the specific mechanisms through which GNNs contribute to the leakage of users' privacy? Does it primarily stem from imbalanced distributions of users' attributes, biased graph structures, or a combination of both factors?

\begin{figure*}
  \centering
  \includegraphics[width=0.8\linewidth]{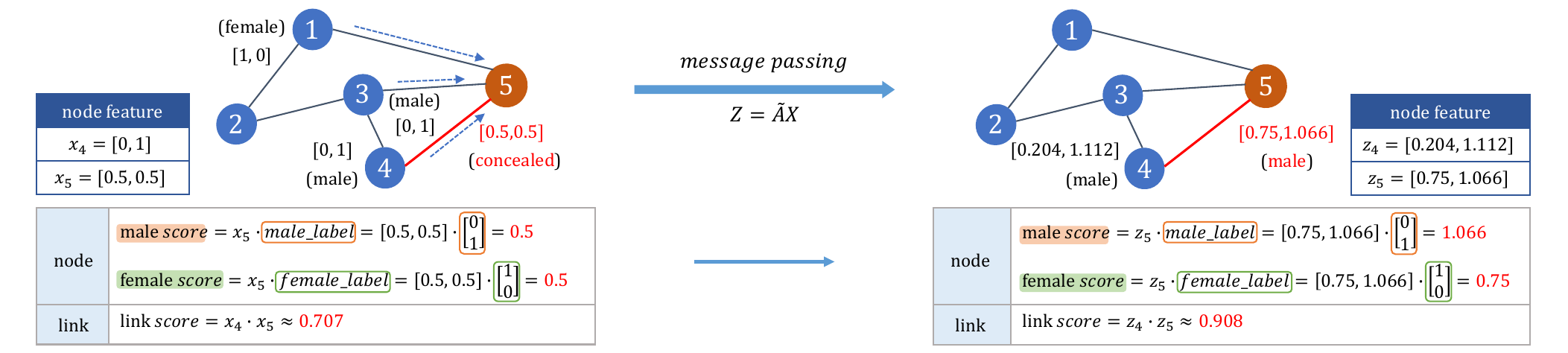}
  \vspace{-10pt}
  \caption{The illustration of dual privacy leakage via message passing under structural bias in GNNs.
  Each node is a user with a binary sensitive attribute \textit{gender}. 
  Users who disclose their gender have features $[1,0]$ (female) or $[0,1]$ (male).
  The target user's (node 5) dual privacy needs to be preserved. For privacy preservation, we can mask his gender information as $[0.5, 0.5]$. The structure is biased as most of target user's neighbors have the same gender. 
   After the message passing, his gender becomes more predictable (node privacy leakage) and his link to user 4 becomes more evident (link privacy leakage).
  In particular, the dot product of his updated feature and the male label increases from 0.5 to 1.066, indicating exacerbated node privacy leakage.
  Further, a simple way for the attacker to infer the link status between users 4 and 5 is to calculate the dot product of user features as belief score. A higher belief score implies a higher confidence in the existence of the link. We observe that, after the GCN-like message passing, the belief score increases ($0.707\rightarrow 0.908$), suggesting an increased risk of link privacy leakage.
  }
  \Description{intro}
  \label{fig:intro}
\end{figure*}

While it may seem intuitive that privacy leakage arises from the divergent distributions of node attributes among different sensitive groups, the precise impact of graph structures on users' privacy leakage remains to be comprehended. GNNs operate through a message-passing scheme, where node embeddings are acquired by aggregating and transforming information from neighboring nodes~\cite{gilmer2017neural}. This mechanism facilitates the propagation of node feature information based on the underlying graph structure, thereby updating node representations across layers. Consequently, it implies that, in addition to the sensitive information present within individual nodes, a biased graph structure has the potential to propagate or even amplify such sensitive information, thereby exacerbating the risks of privacy leakage. We refer to it as privacy under \textit{structural bias}, which is defined as the difference between the number of links that connect nodes with the same and different sensitive attributes (a formal definition is in Def.~\ref{def:sb}). Real-world graphs exhibit inherent structural biases due to phenomena like homophily \cite{zhu2020beyond,khanam2022homophily} and the Matthew effect \cite{merton1988matthew}. In the presence of structural bias, as illustrated in Figure \ref{fig:intro}, we observe that message passing exacerbates the leakage of both node and link privacy, which we refer to as dual privacy leakage. Consequently, this study aims to investigate how message passing leads to privacy leakage and explore methods for preserving dual privacy in GNNs under structural bias.

A potential solution involves a three-stage pipeline approach, starting with debiasing the graph structure and then proceeding with preserving node privacy and link privacy. However, this approach isolates node and link privacy preservation, despite their inherent correlation and their dependence on both node attributes and graph structures. Both privacy-preserving tasks can mutually influence each other and contribute to the debiasing process for graph structures. For instance, to protect users' gender information, it is crucial to identify relevant links that may contribute to privacy leakage, such as connections between users with the same gender information. Additionally, as the objective is to protect dual privacy, finding the optimal privacy-utility trade-off becomes more challenging compared to addressing a single privacy preservation task (either node or link privacy). The desired solution should be able to minimize dual privacy leakage while simultaneously maintaining a competitive level of utility in downstream tasks.

To tackle these challenges, this study begins by conducting a comprehensive theoretical analysis and simulation studies to explore how privacy leakage is propagated and amplified through message passing, particularly under structural bias. Building upon these insights, a principled approach coined \underline{D}ual \underline{P}rivacy-\underline{P}reserving \underline{GNN} (DPPGNN) is proposed, consisting of three key modules. The first module is the Sensitive Information Obfuscation Module (SIO), which aims to protect node-level sensitive attributes by minimizing the distribution distances between node representations of different sensitive groups. The second module is the Dynamic Structure Debiasing Module (DSD), which dynamically debiases the graph structure to mitigate dual privacy leakage. This module aims to alleviate privacy risks associated with biased connections in the graph. The third module is the Adversarial Learning Module (ALM), which employs a min-max game between the graph encoder and an adversary. This module optimizes the trade-off between utility and privacy by iteratively refining the model's ability to preserve privacy while maintaining satisfactory performance in downstream tasks. DPPGNN provides a comprehensive framework for preserving dual privacy in GNNs, addressing the challenges posed by structural bias and the interplay between node and link privacy.

In summary, our main contributions are:
\begin{itemize}
\item \textbf{Problem}: We introduce a novel problem of dual-privacy preservation in GNNs under structural bias. We formulate the problem and formally define structural bias and measure both node and link privacy leakage during the message-passing process. 
\item \textbf{Framework}: We provide the first theoretical analysis and simulation studies to show that message passing propagates and amplifies both node and link privacy leakage under a biased graph structure. We further propose a principled privacy-preserving GNN framework that aims to debias structure and together protect node and link privacy while maintaining competitive performance in downstream tasks. 
\item \textbf{Evaluation}: We compare DPPGNN with the state-of-the-art privacy-preserving GNNs (SOTA) on four benchmark datasets. The experimental results show that DPPGNN shows strong dual privacy-preserving capability and competitive model utility.
\end{itemize}

\section{Notations}
We first introduce the basic notations used throughout the rest of the paper. Let $\mathcal{G}=(\mathcal{V, E})$ denote an undirected attributed graph, where $\mathcal{V}=\{v_1,...,v_{N}\}$ is the set of $N$ nodes and $\mathcal{E}\subseteq N\times N$ is the set of edges.
We use $\mathbf{X}\in \mathbb{R}^{N\times D}$ to denote node attribute matrix, where $\mathbf{x}_i\in \mathbb{R}^D$ denotes the $D-$dimensional attributes of the $i$-th node. $\mathbf{A}\in \mathbb{R}^{N\times N}$ is the adjacency matrix, where $\mathbf{A}_{ij}=1$ if an edge exits between node $v_i$ and $v_j$; otherwise $\mathbf{A}_{ij}=0$.
$\mathbf{D}$ denotes the degree matrix of graphs, which is used for calculating the normalized adjacency matrix $\tilde{\mathbf{A}}$.
$\mathbf{Y}_{u}=\{y_{u,1},...,y_{u,N}\}$ and $\mathbf{Y}_{s}=\{y_{s,1},...,y_{s,N}\}$ denote node utility ($u$) labels and sensitive attributes ($s$), respectively.
Utility label set and sensitive label set are denoted as $\mathcal{Y}_u=\{1,..,C_u\}$ and $\mathcal{Y}_s=\{1,..,C_s\}$, where $C_u$ and $C_s$ represent the number of classes of utility attribute and sensitive attribute, respectively.
A node embedding function $h_{\phi}$ maps input node features $\mathbf{X}$ to latent representations $\mathbf{Z}\in \mathbb{R}^{N\times M}$, where $\mathbf{z}_i=h(\mathbf{x}_i, \mathbf{A})$ is the $i$-th node's embedding and $M$ is the dimension of each node's latent embedding.
$f_{\psi}:\mathbb{R}^M\to \mathbb{R}$ denotes the function used for approximating Wasserstein distance, parameterized by $\psi$. 
$g_{\theta}:\mathbb{R}^M \to \mathbb{R}^{C_u}$ denotes the function used for predicting utility attributes from latent embeddings $\mathbf{Z}$, parameterized by $\theta$.
$g_{\omega}:\mathbb{R}^M \to \mathbb{R}^{C_s}$ denotes the function used for predicting sensitive attributes from $\mathbf{Z}$, parameterized by $\omega$.

\section{Message-Passing in Privacy Leakage}
\label{subsec:PA}
In this section, we provide the first theoretical analysis and simulation studies to show that under structural bias, the message-passing scheme will propagate privacy leakage on graphs, and particularly amplify both node and link privacy leakage.

\subsection{Problem Setup}
Different from conventional privacy-preserving machine learning where samples are independent, the nodes are connected in a graph and their relationships can introduce structural biases due to graph properties such as homophily and the Matthew effect.
We start with a formal definition of structural bias on graphs:
\begin{definition}[Structural Bias]
    Given a graph $\mathcal{G}=(\mathcal{V, E})$ and sensitive attribute $s$, structural bias $B$ is defined as follows:
    \begin{equation}
        B = \frac{1}{|\mathcal{V}|}\sum_{v\in\mathcal{V}}
        \left\vert\frac{|\{u|u\in\mathcal{N}_v, s_u=s_v\}| - |\{u|u\in\mathcal{N}_v, s_u\ne s_v\}|}{|\mathcal{N}_v|} \right\vert,
        \label{def:sb}
    \end{equation}
    where $\mathcal{N}_v$ is the set of neighbors of node $v$. A larger $B\in[0,1]$ implies a more biased graph structure.
\end{definition}
\noindent Based on this definition, structural bias measures the disparity between intra-links and inter-links which are defined w.r.t. sensitive attribute as follows:
\begin{definition}
Given the sensitive attribute $s$ and a link $e$ connecting two nodes $u$ and $v$:
$\mathrm{(1)}$ $e$ belongs to \textbf{intra-links} if $s_u=s_v$;
$\mathrm{(2)}$ $e$ belongs to \textbf{inter-links} if $s_u\ne s_v$.
\end{definition}

We also need to measure the privacy leakage during the GNN graph learning process. Intuitively, when distributions of different sensitive groups are farther apart, it is easier for an attacker (e.g., a sensitive attribute predictor or link predictor) to infer users' sensitive attributes and link status as users from different sensitive groups have more distinct features. Formally, we propose to measure privacy leakage on graphs as
\begin{definition}[Measurement of Privacy Leakage on Graphs]\label{def:pl}
Given the binary sensitive attribute vector $s$, node representations of sensitive groups $s_0$ and $s_1$ follow distributions $\mathbb{P}_0$ and $\mathbb{P}_1$,  respectively, privacy leakage is measured by the distance between $\mathbb{P}_0$ and $\mathbb{P}_1$.
\end{definition}

Note that this definition can be generalized to multi-class sensitive attributes. The majority of privacy-preserving GNNs use inference accuracy of sensitive attributes and links as proxies to quantify privacy leakage. However, these empirical metrics, which will be used in our simulation study in Sec. \ref{simulation}, are incapable to measure the privacy leakage before and after the process of message passing in GNNs. To our knowledge, this is the first attempt to quantify the privacy leakage of message passing during the graph learning process. Other measurements will be explored in the future. 

\begin{figure}
  \centering
  \includegraphics[width=\linewidth]{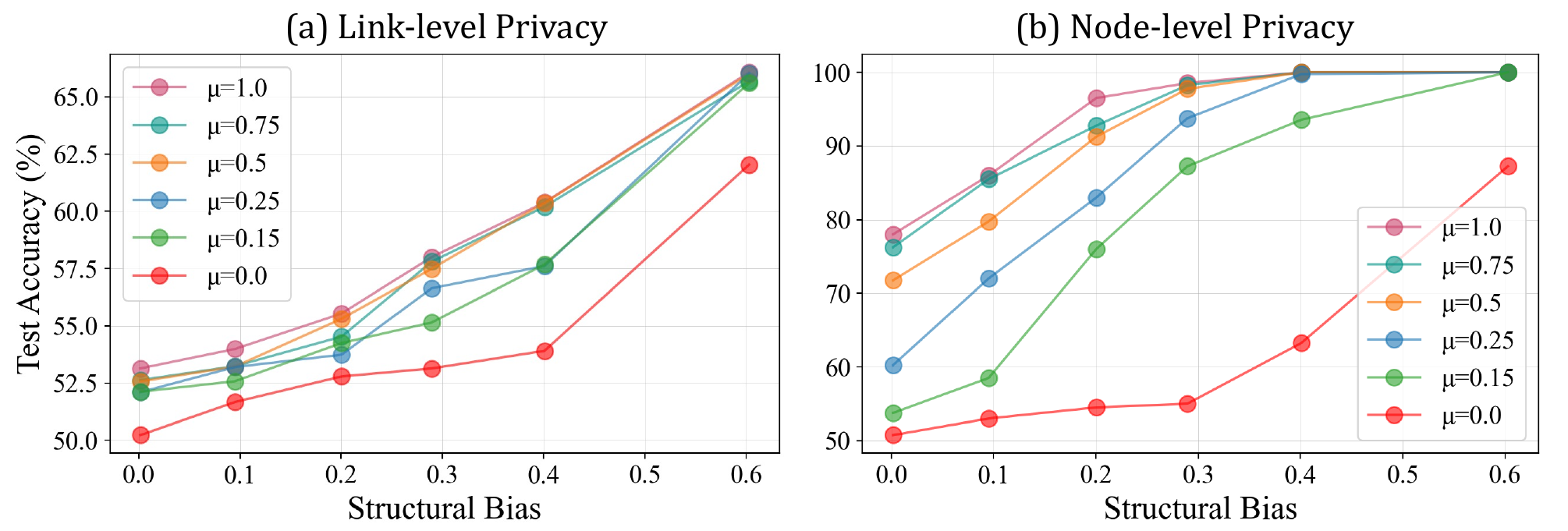}
  \vspace{-20pt}
  \caption{Simulation results. A larger accuracy indicates a more severe node and link privacy leakage.}
  \Description{problem analysis}
  \label{fig:simulation}
\end{figure}

\subsection{Theoretical Analysis}
\label{theory}
To enable the theoretical analysis, we use the random graphs generative framework proposed in \cite{deshpande2018contextual}. From a mathematical perspective, random graphs are often used to answer questions about the properties of typical real-world graphs \cite{janson2011random}. 
We consider graphs $\mathcal{G}^{'}\sim(n, p, q, k, \mu)$ with $n$ nodes, each having a $k$-dimensional feature vector $\mathbf{x}$ and a binary sensitive attribute $s$. The feature vectors are sampled differently depending on $s$: for nodes with sensitive attribute $s_0$ (i.e., $s=0$), each dimension of $\mathbf{x}$ follows $\mathcal{N}(-\mu, 1)$, whereas for nodes with $s=1$, each dimension of $\mathbf{x}$ follows $\mathcal{N}(\mu, 1)$.
To integrate a biased graph structure in $\mathcal{G}^{'}$, we construct edges using two kinds of probabilities: an intra-link probability $p$ and an inter-link probability $q$. Specifically, if the sensitive attributes of two nodes have the same value, an edge between them is generated with the probability $p$, otherwise, with the probability $q$.

Specifically, given a graph $\mathcal{G}^{'}\sim(n, p, q, k, \mu)$, node features for each sensitive group follow the Normal distribution as follows:
\begin{equation}
    \mathbf{x}_0 \sim \mathcal{N}(\mu_{x_0}, \Sigma), \enspace
\mathbf{x}_1 \sim \mathcal{N}(\mu_{x_1}, \Sigma).
\end{equation}
where $\Sigma\in\mathbb{R}^{k\times k}$ denotes the covariance matrix and $\Sigma[i,i]=1$; $\mu_{x_0}, \mu_{x_1}\in\mathbb{R}^{k}$ with $\mu_{x_0}[i] = -\mu, \mu_{x_1}[i] = \mu$$\quad(0\leq i <k)$.
Without loss of generality, Bhattacharyya distance~\cite{bhattacharyya1946measure,bhattacharyya1943measure} is adopted as the metric for the distribution distance measurement in this section for convenience. Lemma~\ref{lemma:11} shows its application in the case of multivariate normal distribution.
\begin{lemma}\label{lemma:11}
Given two multivariate normal distribution 
$p_1\sim\mathcal{N}(\mathbf{\mu}_1, \Sigma_1)$ and
$p_2\sim\mathcal{N}(\mathbf{\mu}_2, \Sigma_2)$, the Bhattacharyya distance between $p_1$ and $p_2$ is given as
\begin{equation*}
    D_B(p_1, p_2) = \frac{1}{8}(\mathbf{\mu}_1 - \mathbf{\mu}_2)^T \Sigma^{-1}(\mathbf{\mu}_1 - \mathbf{\mu}_2) + \frac{1}{2}\ln{ (\frac{\mathrm{det}\Sigma}{\sqrt{\mathrm{det}\Sigma_1\mathrm{det}\Sigma_2}})},
\end{equation*}
where $\Sigma = \frac{\Sigma_1+\Sigma_2}{2}$.
\end{lemma}

Following Definition~\ref{def:pl}, 
the privacy leakage (\textit{PL}) before propagation in GNNs can be quantified as:
\begin{equation}
    PL = D_B(\mathbf{x}_0, \mathbf{x}_1) = \frac{1}{8}\cdot \sum^k(2\mu)^2 = \frac{k}{2}\mu^2.
\end{equation}

Considering a standard message passing $\mathbf{Z}=\tilde{\mathbf{A}}\mathbf{X}$  where $\tilde{\mathbf{A}}$ is the normalized adjacency matrix with self-loop \cite{kipf2016semi},
the representation of node $u$ after propagation can be written as:
\begin{equation}
\begin{aligned}
    z_u
    = \frac{1}{|\mathcal{N}_u|} x_u 
    + \sum_{v\in\mathcal{N}_u, s_u=s_v} \frac{1}{\sqrt{|\mathcal{N}_u||\mathcal{N}_v|}} x_v 
    + \sum_{v\in\mathcal{N}_u, s_u\ne s_v} \frac{1}{\sqrt{|\mathcal{N}_u||\mathcal{N}_v|}} x_v,
\end{aligned}
\end{equation}
where $\mathcal{N}_u$ denotes the set of neighboring nodes of node $u$, $s_u$ denotes the sensitive group of node $u$.
Thus node representations of each sensitive group after the propagation follow distributions:
\begin{equation}
\begin{aligned}
    \mathbf{z}_0\sim\mathcal{N}(\mu_{z_0}, \Sigma_{z_0}),\enspace
    \mathbf{z}_1\sim\mathcal{N}(\mu_{z_1}, \Sigma_{z_1}),
\end{aligned}
\end{equation}
where
\begin{equation*}
\begin{aligned}
\mu_{z_0}[i] &=-  \frac{\mu}{|\mathcal{N}_u|} - \sum_{v\in\mathcal{N}_u, s_u=s_v} \frac{\mu}{\sqrt{|\mathcal{N}_u||\mathcal{N}_v|}} +\sum_{v\in\mathcal{N}_u, s_u\ne s_v} \frac{\mu}{\sqrt{|\mathcal{N}_u||\mathcal{N}_v|}}, \\
\mu_{z_1}[i] &=  \frac{\mu}{|\mathcal{N}_u|} + \sum_{v\in\mathcal{N}_u, s_u=s_v} \frac{\mu}{\sqrt{|\mathcal{N}_u||\mathcal{N}_v|}} -\sum_{v\in\mathcal{N}_u, s_u\ne s_v} \frac{\mu}{\sqrt{|\mathcal{N}_u||\mathcal{N}_v|}},
\end{aligned}
\end{equation*}
and
\begin{equation*}
 \Sigma_{z_0}[i,i] = \Sigma_{z_1}[i,i] = \frac{1}{|\mathcal{N}_u|^2}+\sum_{v\in\mathcal{N}_u}\frac{1}{|\mathcal{N}_u||\mathcal{N}_v|} \quad( 0\leq i<k).
\end{equation*}

Considering the generation process of synthetic graphs, for each node, the size of neighboring nodes set could be approximated as $n(p+q)$; a proportion of $\frac{p}{p+q}$ of its neighbors shares the same sensitive group as itself, and a proportion of $\frac{q}{p+q}$ of neighbors belongs to the opposite sensitive group.
Thus $\mu_{z_0}$ and $\mu_{z_1}$ could be further written as:
\begin{equation}
\begin{aligned}
    \mu_{z_0}[i] = \frac{n(q-p)-1}{n(p+q)}\cdot\mu,\enspace
    \mu_{z_1}[i] = \frac{n(p-q)+1}{n(p+q)}\cdot\mu.
\end{aligned}
\end{equation}

Similarly, we could simplify $\Sigma_{z_0}$ and $\Sigma_{z_1}$ as:
\begin{equation}
 \Sigma_{z_0}[i,i] = \Sigma_{z_1}[i,i] = \frac{1+n(p+q)}{n^2(p+q)^2}
\end{equation}
where $0\leq i<k$.
Using Lemma~\ref{lemma:11}, privacy leakage after message passing can be quantified as:
\begin{equation}
\begin{aligned}
    PL' = D_B(\mathbf{z}_0, \mathbf{z}_1) 
    = \frac{k}{2}\mu^2\cdot\frac{n(p+q)(np-nq+1)^2}{1+n(p+q)}.
\end{aligned}
\end{equation}


We can then have the following proposition:
\begin{proposition}
\label{pp1}
Given a graph $\mathcal{G}'\sim (n,p,q,k,\mu)$,
\\
$\mathrm{(1)}$ privacy leakage in GNNs is exacerbated after a standard GCN-like message passing if the structural bias $B$ is above a threshold, i.e., 
\begin{equation*}
\small
    \Delta PL=(PL'-PL)>0, \text{ if } B>\frac{\sqrt{n^2(p+q)^2+n(p+q)}-n(p+q)}{n^2(p+q)^2}.
\end{equation*}
\\
$\mathrm{(2)}$ when $\mu$ and $k$ are fixed, a larger structural bias $B$ leads to more severe privacy leakage $\Delta PL$, i.e.,
\begin{equation*}
    \text{ If } B_1>B_2, \text{ then } \Delta PL_1>\Delta PL_2.
\end{equation*}

\end{proposition}

The detailed proof is given in Appendix~\ref{appdix:1}.
Proposition~\ref{pp1} suggests that (1) the message-passing scheme amplifies privacy leakage when the structural bias exceeds a certain threshold; and (2) after exceeding the threshold, 
privacy leakage grows with increasing structural bias.
This result implies that message-passing process in GNNs propagates and embeds sensitive information into node representations. Attackers can then infer sensitive attributes and the link status from the learned node representations.

\subsection{Simulation Study}
\label{simulation}
We further conduct a series of simulation studies to support our theoretical analysis. We measure node-level and link-level privacy leakage using standard metrics: sensitive attributes inference accuracy and link prediction accuracy. A larger accuracy indicates a more severe privacy leakage. For synthetic graphs generation, we set $n=1000$, $k=4$, and vary the values of $p$, $q$, $\mu$. Specifically, to generate graphs with different levels of structural bias, we vary $(p,q)$ among $\{$(0.08, 0.02), (0.07, 0.03), (0.065, 0.035), (0.06, 0.04), (0.055, 0.045),(0.05, 0.05)$\}$, and a larger $|p-q|$ yields graphs with larger structural bias. We also examine how the distance between distributions of node attributes for two sensitive groups influences privacy disclosure on graphs. In particular, we vary $\mu$ among $\{$0, 0.15, 0.25, 0.5, 0.75, 1$\}$, and a greater $\mu$ indicates a larger distribution distance between two groups. 

We use a one-layer GCN combined with a multilayer perceptron as the threat model (i.e., attackers) which is then trained on 60\% of nodes (i.e., training set) whose sensitive attributes are visible to adversaries. 
Similarly, the threat model for link privacy is trained on 60\% of links (i.e., training set) on the graph visible to the adversary.

Simulation results are shown in Fig.~\ref{fig:simulation}. 
We observe that (1) a more biased graph structure leads to both greater node and link privacy leakage. In addition, privacy leakage w.r.t. link consistently increases as bias enlarges while that w.r.t. node quickly reaches the largest. (2) When the structural bias is fixed, with larger values of $\mu$, i.e., the feature distributions for different sensitive groups are more distant, both node and link privacy leakage get more severe after the GNN learning procedure. This implies that message passing in GNNs can propagate and amplify sensitive information via graph structure, and our privacy leakage measurement in Def. \ref{def:pl} is a proper indicator of node and link privacy disclosure. Both observations support the theoretical analysis in Sec. \ref{theory}.
\newline

\noindent\textbf{Summary. }Both theoretical analysis and simulation studies based on random graphs suggest that message passing plays a critical role in the propagation of sensitive information in GNNs and a biased structure can exacerbate both node and link-level privacy leakage. Therefore, it is important to preserve dual privacy while accounting for structural bias given the ubiquity of such biased graph structures and the wide applications of GNNs in critical domains. 

\begin{figure}
  \centering
  \includegraphics[width=\linewidth]{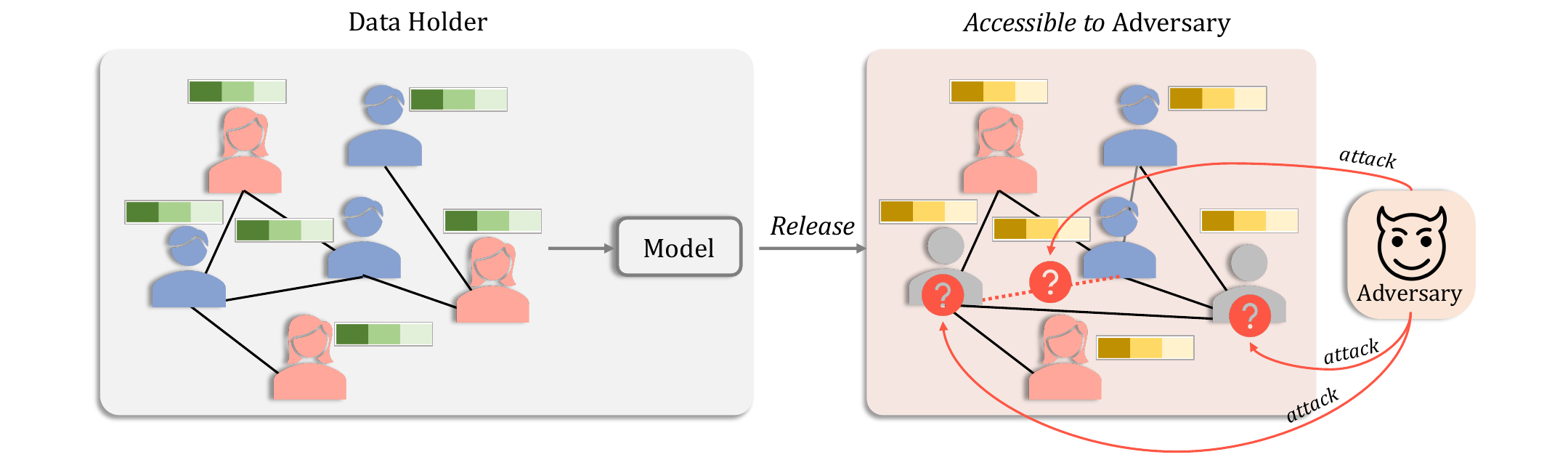}
  \vspace{-25pt}
  \caption{Illustration of the attacking setting.}
  \Description{Attack Setting}
  \label{fig:setting}
\end{figure}
\section{Problem Formulation}
With the amplified dual privacy leakage, this work aims to learn dual privacy-preserving node representations and a debiased graph structure in GNNs while maximizing the useful information reserved for downstream tasks such as node classification. 
\begin{definition}[Dual-Privacy Preservation under Structural Bias] Given a graph $\mathcal{G=(V,E)}$ with a biased adjacency matrix $\mathbf{A}$ and feature matrix $\mathbf{X}$, we aim to learn a privacy-preserving embedding function $h_{\phi}: (\mathbf{X},\mathbf{A}) \rightarrow \mathbf{Z}$ and adjacency matrix $\tilde{\mathbf{A}}$ to achieve the following two goals: \textcircled{1} an adversary cannot easily infer node sensitive attribute and link status from $\mathbf{Z}$ and $\tilde{\mathbf{A}}$; \textcircled{2} $\mathbf{Z}$ and $\tilde{\mathbf{A}}$ retain most of the information useful for the downstream task. \end{definition}

Note that this work adopts a black-box attack setting (Fig.~\ref{fig:setting}), i.e., the adversary can only access the node embeddings learned via the graph encoder and has no knowledge of the encoder's parameters.

\section{The Proposed Framework}
This section details the design of the proposed framework -- \underline{D}ual-\underline{P}rivacy \underline{P}reserving \underline{GNN}s (DPPGNN). We start with an overview of DPPGNN, followed by introductions to each individual module. 

\begin{figure*}
  \centering
  \includegraphics[width=0.8\linewidth]{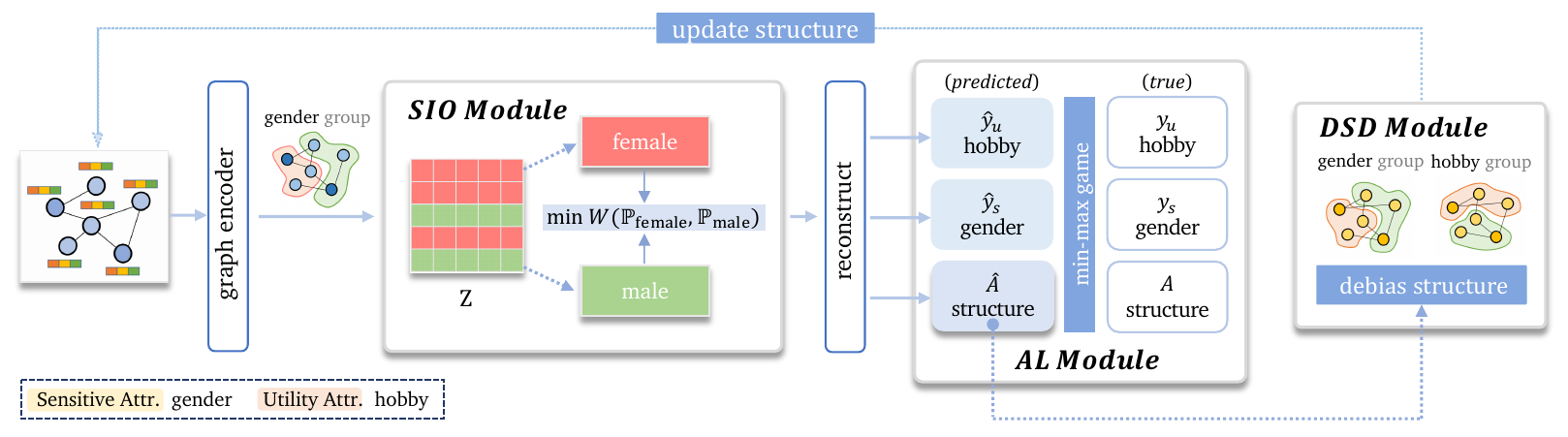}
  \vspace{-15pt}
  \caption{DPPGNN includes three modules: Sensitive Information Obfuscation (SIO), Dynamic Structure Debiasing (DSD), and Adversarial Learning Module (ALM).
  We use gender and hobby as the sensitive and utility attributes, respectively.}
  \Description{the overall framework.}
  \label{fig:framework}
\end{figure*}
\subsection{Overview}
Both the theoretical analysis and simulation study in Sec. \ref{subsec:PA} suggest that dual-privacy preservation on GNNs under structural bias is critical.
Major challenges are from two aspects: (1) How to alleviate graph structure bias that exacerbates dual privacy issues in GNNs? Meanwhile, (2) how to achieve a more optimal privacy-utility trade-off when both node and link privacy shall be guaranteed? 

To overcome the \textbf{first challenge}, we first propose to preserve privacy on graphs by approximating and obfuscating distributions of node representations for different sensitive groups, and then \textit{dynamically} identify the biased structure and reduce their significance. This can explicitly reduce the structural bias and preserve dual privacy.
To overcome the \textbf{second challenge}, we design an adversarial learning framework to optimize the utility-privacy trade-off, striking a balance between preserving privacy and maintaining competitive utility. In particular, DPPGNN includes the following three modules: Sensitive Information Obfuscation (SIO) module aims to obfuscate sensitive information of node representations learned by the graph encoder. Dynamic Structure Debiasing (DSD) module dynamically debiases graph structure by reducing the importance of biased structures defined based on sensitive groups. Its output is a debiased adjacency matrix. Adversarial Learning Module (ALM) takes input from SIO and plays a min-max game between the graph encoder and an adversary in an adversarial environment to achieve a better utility-privacy trade-off.

Note that while we modularize the framework, each component in DPPGNN is intertwined to together learn privacy-preserving node representations and debias graph structures. The overall framework is illustrated in Fig. \ref{fig:framework}.

\subsection{Sensitive Information Obfuscation}
The SIO Module aims to obfuscate sensitive information embedded in the learned node representations $\mathbf{Z}$. The basic idea is that the risk of node privacy leakage increases when distributions of different sensitive groups are distinguishable and can be easily differentiated by an attacker. Therefore, SIO first estimates the distributions for each sensitive group based on the learned node embeddings and then minimizes their distance.

Following \cite{arjovsky2017wasserstein}, we employ Wasserstein distance and approximate the distribution distance between different sensitive groups based on their node representations $\mathbf{Z}$.
Let $\mathbb{P}_j$ and  $\mathbb{P}_k$ denote the feature distributions of sensitive group $j$ and $k$ respectively. $\mathbf{z}_{j}$ and $\mathbf{z}_{k}$ are two random variables sampled from $\mathbb{P}_j$ and $\mathbb{P}_k$ separately, 
then the Wasserstein-1 distance $D_W(\cdot)$ between $\mathbb{P}_j$ and  $\mathbb{P}_k$ is defined as: 
\begin{equation}
\label{eq:1}
    D_W(\mathbb{P}_j, \mathbb{P}_k) = \inf_{\gamma \in \prod (\mathbb{P}_j, \mathbb{P}_k)} \mathbb{E}_{(\mathbf{z}_{j},\mathbf{z}_{k})\sim \gamma} [||\mathbf{z}_{j}-\mathbf{z}_{k}||],
\end{equation}
where $\prod (\mathbb{P}_j, \mathbb{P}_k)$ is the set of all joint distributions $\gamma (\mathbf{z}_j, \mathbf{z}_k)$ whose marginals are $\mathbb{P}_j$ and $\mathbb{P}_k$, respectively.


The duality theorem of Kantorovich and Rubinstein \cite{villani2009optimal} allows us to transform the intractable infimum in Eq. (\ref{eq:1}) into:
\begin{equation}
    D_W(\mathbb{P}_j, \mathbb{P}_k) = \frac{1}{K}\sup_{||f||_L \le K} \mathbb{E}_{\mathbf{z}_{j}\sim \mathbb{P}_j}[f(\mathbf{z}_{j})] - \mathbb{E}_{\mathbf{z}_{k}\sim \mathbb{P}_k} [f(\mathbf{z}_{k})],
\end{equation}
where the $\sup$ is over all the $K$-Lipschitz functions $f:\mathbb{R}^{M}\to \mathbb{R}$. 
In practice, $f(\cdot)$ is usually parameterized as a neural network with weights $\psi$, which are subject to $K$-Lipschitz constraint $||f_{\psi}||_L \le K$. To enforce the constraint, weights are clipped into a fixed range after each update. 
Approximated $D_W$ can eventually be obtained via solving the following problem
\begin{equation}\label{eq:wd}
    \max_{\psi} \mathbb{E}_{\mathbf{z}_{j}\sim \mathbb{P}_j}[f_{\psi}(\mathbf{z}_{j})] - \mathbb{E}_{\mathbf{z}_{k}\sim \mathbb{P}_k} [f_{\psi}(\mathbf{z}_{k})].
\end{equation}

Our goal is to achieve information obfuscation by minimizing the distribution distance between different sensitive groups. Accordingly,
we formulate the objective function for the SIO module as follows:
\begin{equation}
\begin{aligned}
     \min_{\psi}\mathcal{L}_{SIO} = \min_{\psi} (\sum_{j} \sum_{k\ne j} \mathbb{E}_{\mathbf{z}_{j}\sim \mathbb{P}_j}[f_{\psi}(\mathbf{z}_{j})] - \mathbb{E}_{\mathbf{z}_{k}\sim \mathbb{P}_k} [f_{\psi}(\mathbf{z}_{k})])& \\ \forall j, k\in \mathcal{Y}_s.&
\end{aligned}
\end{equation}




\subsection{Dynamic Structure Debiasing}
As GNNs amplify dual privacy leakage via message passing under biased structures, we propose the DSD Module to dynamically refine the adjacency matrix to mitigate structural bias. 

Based on Def. \ref{def:sb}, the graph structure becomes biased when the intra-links w.r.t. sensitive attribute (i.e., links connecting nodes from the same sensitive group) outnumber inter-links (i.e., links connecting nodes from different sensitive groups).
Thus, to mitigate this bias, we should reduce the importance of these intra-links.
We propose to update the graph structure during training such that the number of intra- and inter-links is balanced and $B$ is minimized.
It should be noted that intra-links w.r.t. utility attributes 
are useful for the downstream task. Therefore, when minimizing structural bias to preserve privacy, we should also maintain sufficient information on intra-links w.r.t. model utility.

Specifically, we first reconstruct the graph structure $\widehat{A}$ from the learned node embeddings as follows:
\begin{equation}
    \widehat{A} = \sigma (\mathbf{Z}\mathbf{Z}^T),
\end{equation}
where $\sigma(\cdot)$ represents the sigmoid function. 
The reconstructed adjacency matrix reflects the biased structural information that the node embeddings capture and that the adversary can exploit.
Next, we minimize the difference between the numbers of intra-links and inter-links w.r.t. sensitive attribute using Eq. (\ref{obj:adj}) with $\widehat{A}$:

\begin{equation}
\label{obj:adj}
    \mathcal{L}_{adj}^{(s)} = ||S_{intra}^{(s)} - S_{inter}^{(s)}||^2,
\end{equation}
where

\begin{equation}
\label{eq:intra}
\begin{aligned}
    &S_{intra}^{(s)} = \sum_{j,k\in\mathcal{V}, y_{s,j}=y_{s,k}} \widehat{A}_{j,k},\enspace
    S_{inter}^{(s)} = \sum_{j,k\in\mathcal{V}, y_{s,j}\ne y_{s,k}} \widehat{A}_{j,k}.
\end{aligned}
\end{equation}
$y_{s,j}$ and $y_{s,k}$ denote the sensitive labels of node $j$ and $k$, respectively.

We can obtain a similar loss function w.r.t. model utility label as:


\begin{equation}
\label{eq:adj}
    \mathcal{L}_{adj}^{(u)} = ||S_{intra}^{(u)} - S_{inter}^{(u)}||^2,
\end{equation}
where

\begin{equation}
\label{eq:inter}
\begin{aligned}
    S_{intra}^{(u)} = \sum_{j,k\in\mathcal{V}, y_{u,j}=y_{u,k}} \widehat{A}_{j,k},\enspace S_{inter}^{(u)} = \sum_{j,k\in\mathcal{V}, y_{u,j}\ne y_{u,k}} \widehat{A}_{j,k}.
\end{aligned}
\end{equation}
$y_{u,j}$ and $y_{u,k}$ denote the utility labels of node $j$ and $k$, respectively.

The goal of DSD is to
(1) minimize $\mathcal{L}_{adj}^{(s)}$ to debias structure; and
(2) maximize $\mathcal{L}_{adj}^{(u)}$ to improve model utility.
Therefore, combining Eq. (\ref{obj:adj}) and Eq. (\ref{eq:adj}), the final objective function is defined as follows:
\begin{equation}
\label{obj:dsd}
    \min_{\tilde{A}}\mathcal{L}_{DSD} = \min_{\tilde{A}} (\alpha \mathcal{L}_{adj}^{(s)} - \mathcal{L}_{adj}^{(u)}),
\end{equation}
where $\alpha$ is the hyperparameter used to balance different terms.

To update the adjacency matrix based on Eq. (\ref{obj:dsd}), we employ the optimization strategy in \cite{li2021dyadic}.
To be specific, let $\tilde{A}$ denote the normalized adjacency matrix, which is initialized via left normalization $\tilde{A}=\mathbf{D}^{-1}A$.
In each iteration, $\tilde{A}$ is updated by $\tilde{A}_{new} = \tilde{A}_{old}-\eta \nabla \mathcal{L}_{DSD}$. To maintain training stability, the updated $\tilde{A}$ should be a right stochastic matrix. The projected gradient descent (PGD) algorithm proposed in \cite{wang2013projection} is then applied to satisfy this condition.

\subsection{Improving Privacy-Utility Trade-off}
One major challenge in dual-privacy preservation is a more  severe privacy-utility trade-off than the standard single-privacy preservation. The ALM aims to achieve a better trade-off via adversarial learning, which plays a min-max game between the graph encoder and an adversary in the training process. It has been shown in multiple studies \cite{xiao2020adversarial,tripathy2019privacy} that adversarially trained neural networks can attain optimal privacy-utility trade-off.
 

In adversarial learning for privacy preservation, an adversary tries to infer the sensitive attributes and link status from the accessible node embeddings $\mathbf{Z}$ by minimizing the following loss:
\begin{equation}
\begin{aligned}
    &\min_{\omega}\mathcal{L}_{privacy} = \min_{\omega}(\mathcal{L}_{priv\_attr} + \mu\mathcal{L}_{link} )\\
    &= \min_{\omega}(\sum_{v_i\in \mathcal{V}} l(g_{\omega}(\mathbf{z}_i), y_{s, i})
    + \mu\sum_{v_i, v_j \in \mathcal{V}} l(\tilde{A}_{ij}, A_{ij})),
\end{aligned}
\end{equation}
where $g_{\omega}: \mathbb{R}^{M}\to \mathbb{R}^{|\mathcal{Y}_s|}$ is
a multi-layer perceptron (MLP) that predicts the sensitive attributes given node embeddings, with $\omega$ as its parameters;
$l(\cdot)$ is a loss function such as cross-entropy loss; $\mu$ is a hyperparameter controlling the contribution of each term.

On the other hand, ALM seeks to guarantee model utility by minimizing the following objective function:
\begin{equation}
    \min_{\phi,\theta}\mathcal{L}_{utility} = \min_{\phi,\theta}
    \sum_{v_i\in \mathcal{V}} l(g_{\theta}(\mathbf{z}_i), y_{u, i}),
\end{equation}
where function $g_{\theta}: \mathbb{R}^{M}\to \mathbb{R}^{|\mathcal{Y}_u|}$ is a MLP parameterized by $\theta$ that predicts utility labels of nodes.

The overall objective function for ALM can be written as:
\begin{equation}
    \min_{\omega,\phi,\theta} \mathcal{L}_{ALM} = \min_{\omega,\phi,\theta} (\mathcal{L}_{utility} - \lambda\mathcal{L}_{privacy}).
\end{equation}
We alternatively update the encoder ($\mathcal{L}_{utility}$) and adversary ($\mathcal{L}_{privacy}$) to remove sensitive information from the node embeddings $\mathbf{Z}$ and improve the privacy-utility trade-off.

\noindent\textbf{Overall Algorithm.} In summary, the overall objective function of DPPGNN is given by:
\begin{equation}
    \min\mathcal{L} = \min \mathcal{L}_{SIO} + \mathcal{L}_{DSD} + \mathcal{L}_{ALM}.
\end{equation}
We alternatively update each module in each epoch.

\section{Experiments}
In this section, we evaluate the effectiveness of DPPGNN by answering the following three research questions:
\begin{itemize}
\item \textbf{RQ. 1}. How does DPPGNN fare against the state-of-the-art privacy-preserving GNN models for the node and link privacy-preserving tasks?
\item \textbf{RQ. 2}. How does DPPGNN fare against the state-of-the-art privacy-preserving GNN models on downstream tasks such as the node classification task?
\item \textbf{RQ. 3}. How does each module of DPPGNN impact its performance w.r.t. both utility and privacy? And how effective is the structural debiasing function $\mathcal{L}_{DSD}$?
\end{itemize}

To answer \textbf{RQ. 1-RQ. 2}, we compare DPPGNN with seven state-of-the-art baselines for utility measure (i.e., node classification) and privacy measure (i.e., node and link privacy preservation). To answer \textbf{RQ. 3}, we conduct a series of ablation studies to examine the influence of each module and the parameter $\alpha$ in DPPGNN.
The implementation code will be released in due course.

\subsection{Experimental Setup}
This section describes the datasets, baselines, privacy attack settings, parameter settings of DPPGNN, and metrics used for evaluating models' utility and privacy preservation performance. 

\subsubsection{Datasets}

We evaluate DPPGNN on four benchmark datasets, including Rochester38, Yale4 \cite{traud2012social}, German Credit and Recidivism \cite{agarwal2021towards}. 
The German Credit dataset contains data from a German bank with client gender as a sensitive attribute and credit risk level as a utility label. 
Recidivism consists of bail-released defendants with race as a sensitive attribute and bail decision as a utility label. Rochester38 and Yale4 are two Facebook friendship networks. Rochester38 has gender as a sensitive attribute and class year as a utility label, while Yale4 has class year as a sensitive attribute and student status flag as a utility label.
See Table~\ref{tab:dataset} for their basic statistics.
We also measure their structural bias using the metric proposed in Def.~\ref{def:sb}. Results show that German Credit and Recidivism have the largest and least structural bias scores, respectively.

\begin{table}[]
\caption{Statistics of four benchmark datasets.}
\vspace{-8pt}
\label{tab:dataset}
\resizebox{\columnwidth}{!}{%
\begin{tabular}{lcccc}
\toprule
\textbf{Dataset}      & \textbf{German Credit} & \textbf{Rochester38} & \textbf{Yale4} & \textbf{Recidivism}\\
\midrule
\textbf{\#Nodes}      & 1,000    & 4,563      & 8,578  &  18,876      \\
\textbf{\#Edges}      & 22,242  & 167,653     & 405,450  &  321,308    \\
\textbf{\#Attributes} & 27     & 7         & 7  & 18 \\  
\textbf{Structural Bias } & 0.605 & 0.278 & 0.414 & 0.208 \\
\bottomrule
\end{tabular}%
}
\end{table}

\begin{table*}
\caption{Performance of node classification, sensitive attribute inference, and link status inference on three benchmark datasets. $\uparrow$ denotes that a higher value is more desired.}\label{tab:result}
\centering
\vspace{-8pt}
\resizebox{\textwidth}{!}{%
\begin{tabular}{c|c||cccc||cccc||cccc}
\hline
 &
   &
  \multicolumn{4}{c||}{\begin{tabular}[c]{@{}c@{}}\# Utility ($\uparrow$)\\ (node classification performance)\end{tabular}} &
  \multicolumn{4}{c||}{\begin{tabular}[c]{@{}c@{}}\# Node-level Privacy ($\downarrow$)\\ (sensitive attribute inference performance)\end{tabular}} &
  \multicolumn{4}{c}{\begin{tabular}[c]{@{}c@{}}\# Link-level Privacy ($\downarrow$)\\ (link inference performance)\end{tabular}} \\ \cline{2-14} 
Method &
  Dataset &
  Rochester38 &
  Yale4 &
  German Credit &
  Recidivism &
  Rochester38 &
  Yale4 &
  German Credit &
  Recidivism &
  Rochester38 &
  Yale4 &
  German Credit &
  Recidivism \\ \hline
\multirow{2}{*}{GCN} &
  Acc. &
  93.56 $\pm$ 0.16 &
  \underline{96.24 $\pm$ 0.21} &
  72.49 $\pm$ 0.42 &
  93.92 $\pm$ 0.41 &
  55.68 $\pm$ 0.23 &
  79.97 $\pm$ 0.65 &
  78.86 $\pm$ 0.31 &
  55.57 $\pm$ 0.55 &
  82.96 $\pm$ 0.14 &
  81.36 $\pm$ 0.10 &
  76.34 $\pm$ 2.35 &
  79.96 $\pm$ 0.97 \\
 &
  F1 &
  91.53 $\pm$ 0.09 &
  95.52 $\pm$ 0.18 &
  56.87 $\pm$ 0.24 &
  93.45 $\pm$ 0.45 &
  54.41 $\pm$ 0.56 &
  78.23 $\pm$ 1.12 &
  78.60 $\pm$ 0.25 &
  55.39 $\pm$ 0.54 &
  82.47 $\pm$ 0.08 &
  81.00 $\pm$ 0.07 &
  76.26 $\pm$ 2.37 &
  79.86 $\pm$ 1.00 \\ \hline
\multirow{2}{*}{LapGraph} &
  Acc. &
  90.11 $\pm$ 0.57 &
  94.11 $\pm$ 0.29 &
  69.83 $\pm$ 0.09 &
  89.48 $\pm$ 0.02 &
  54.73 $\pm$ 0.31 &
  67.73 $\pm$ 5.24 &
  \textbf{70.89 $\pm$ 0.14} &
  54.65 $\pm$ 0.13 &
  80.63 $\pm$ 0.72 &
  74.53 $\pm$ 3.48 &
  \underline{59.89 $\pm$ 0.48} &
  73.91 $\pm$ 0.08 \\
 &
  F1 &
  88.92 $\pm$ 0.68 &
  93.72 $\pm$ 0.32 &
  46.25 $\pm$ 0.25 &
  88.52 $\pm$ 0.03 &
  53.28 $\pm$ 0.59 &
  67.57 $\pm$ 5.17 &
  \textbf{55.49 $\pm$ 0.40} &
  54.29 $\pm$ 0.18 &
  80.59 $\pm$ 0.74 &
  74.5 $\pm$ 3.47 &
  \underline{59.52 $\pm$ 0.91} &
  73.78 $\pm$ 0.09 \\ \hline
\multirow{2}{*}{PPRLG} &
  Acc. &
  92.11 $\pm$ 1.26 &
  \textbf{96.56 $\pm$ 0.53} &
  73.06 $\pm$ 1.24 &
  93.63 $\pm$ 0.36 &
  55.04 $\pm$ 0.80 &
  78.19 $\pm$ 2.62 &
  76.43 $\pm$ 0.29 &
  54.85 $\pm$ 0.13 &
  80.88 $\pm$ 0.7 &
  81.13 $\pm$ 1.06 &
  68.22 $\pm$ 1.44 &
  68.12 $\pm$ 0.54 \\
 &
  F1 &
  91.08 $\pm$ 1.42 &
  \textbf{96.30 $\pm$ 0.56} &
  59.17 $\pm$ 1.21 &
  93.13 $\pm$ 0.39 &
  53.86 $\pm$ 0.68 &
  77.66 $\pm$ 2.42 &
  65.71 $\pm$ 1.07 &
  54.84 $\pm$ 0.12 &
  80.83 $\pm$ 0.71 &
  81.11 $\pm$ 1.06 &
  68.19 $\pm$ 1.30 &
  67.03 $\pm$ 0.55 \\ \hline
\multirow{2}{*}{APGE} &
  Acc. &
  92.43 $\pm$ 0.17 &
  82.51 $\pm$ 0.32 &
  70.03 $\pm$ 0.31 &
  92.18 $\pm$ 0.27 &
  62.33 $\pm$ 0.45 &
  \underline{57.86 $\pm$ 0.52} &
  \underline{69.63 $\pm$ 0.37} &
  55.70 $\pm$ 0.25 &
  82.72 $\pm$ 0.31 &
  77.69 $\pm$ 0.56 &
  77.06 $\pm$ 0.68 &
  86.72 $\pm$ 0.18 \\
 &
  F1 &
  92.16 $\pm$ 0.12 &
  82.18 $\pm$ 0.33 &
  50.09 $\pm$ 0.69 &
  91.51 $\pm$ 0.36 &
  61.97 $\pm$ 0.50 &
  \underline{57.47 $\pm$ 0.51} &
  \textbf{51.34 $\pm$ 0.47} &
  55.49 $\pm$ 0.17 &
  82.69 $\pm$ 0.29 &
  77.64 $\pm$ 0.54 &
  77.03 $\pm$ 0.71 &
  86.61 $\pm$ 0.16 \\ \hline
\multirow{2}{*}{GAE} &
  Acc. &
  93.86 $\pm$ 0.30 &
  94.44 $\pm$ 0.88 &
  73.30 $\pm$ 0.26 &
  \textbf{95.11 $\pm$ 0.12} &
  55.40 $\pm$ 0.12 &
  76.09 $\pm$ 0.76 &
  73.52 $\pm$ 0.93 &
  53.86 $\pm$ 0.07 &
  72.99 $\pm$ 0.61 &
  81.46 $\pm$ 0.18 &
  79.77 $\pm$ 0.92 &
  76.26 $\pm$ 0.83 \\
 &
  F1 &
  \underline{92.80 $\pm$ 0.34} &
  94.06 $\pm$ 0.95 &
  61.21 $\pm$ 0.94 &
  \textbf{94.75 $\pm$ 0.13} &
  54.29 $\pm$ 0.81 &
  75.88 $\pm$ 0.67 &
  63.43 $\pm$ 0.70 &
  52.72 $\pm$ 0.44 &
  72.86 $\pm$ 0.71 &
  81.42 $\pm$ 0.20 &
  79.71 $\pm$ 0.97 &
  76.00 $\pm$ 0.93 \\ \hline
\multirow{2}{*}{GAE-NL} &
  Acc. &
  \textbf{94.06 $\pm$ 0.29} &
  96.06 $\pm$ 0.30 &
  \textbf{75.27 $\pm$ 0.27} &
  \underline{94.57 $\pm$ 0.28} &
  \underline{53.83 $\pm$ 0.36} &
  65.08 $\pm$ 0.61 &
  72.33 $\pm$ 0.57 &
  53.93 $\pm$ 0.11 &
  72.94 $\pm$ 0.54 &
  72.19 $\pm$ 0.29 &
  70.16 $\pm$ 1.03 &
  69.97 $\pm$ 0.53 \\
 &
  F1 &
  \textbf{92.96 $\pm$ 0.35} &
  95.23 $\pm$ 0.35 &
  \underline{65.93 $\pm$ 0.78} &
  \underline{94.17 $\pm$ 0.30} &
  \underline{51.40 $\pm$ 0.19} &
  64.65 $\pm$ 0.78 &
  59.29 $\pm$ 1.17 &
  52.73 $\pm$ 0.28 &
  72.84 $\pm$ 0.56 &
  72.09 $\pm$ 0.26 &
  70.12 $\pm$ 1.03 &
  69.13 $\pm$ 0.75 \\ \hline
\multirow{2}{*}{DPGCN} &
  Acc. &
  89.85 $\pm$ 0.47 &
  95.98 $\pm$ 0.38 &
  72.97 $\pm$ 1.03 &
  88.72 $\pm$ 0.32 &
  53.95 $\pm$ 0.67 &
  61.27 $\pm$ 0.17 &
  75.90 $\pm$ 0.83 &
  \underline{53.37 $\pm$ 0.12} &
  \underline{71.57 $\pm$ 0.19} &
  \underline{71.72 $\pm$ 0.91} &
  67.44 $\pm$ 0.45 &
  \underline{67.49 $\pm$ 0.31} \\
 &
  F1 &
  86.60 $\pm$ 0.45 &
  \underline{95.83 $\pm$ 0.37} &
  59.41 $\pm$ 0.57 &
  87.56 $\pm$ 0.38 &
  51.99 $\pm$ 0.93 &
  61.90 $\pm$ 0.18 &
  66.76 $\pm$ 0.73 &
  \underline{52.61 $\pm$ 0.31} &
  \underline{71.53 $\pm$ 0.13} &
  \underline{71.69 $\pm$ 0.90} &
  67.35 $\pm$ 0.43 &
  \underline{67.42 $\pm$ 0.21} \\ \hline
\multirow{2}{*}{\textbf{DPPGNN(ours)}} &
  Acc. &
  \underline{93.89 $\pm$ 0.32} &
  95.20 $\pm$ 0.81 &
  \underline{73.32 $\pm$ 0.12} &
  92.28 $\pm$ 0.03 &
  \textbf{52.08 $\pm$ 0.47} &
  \textbf{54.88 $\pm$ 0.70} &
  \textbf{69.46 $\pm$ 0.33} &
  \textbf{52.07 $\pm$ 0.80} &
  \textbf{63.78 $\pm$ 0.99} &
  \textbf{70.83 $\pm$ 0.69} &
  \textbf{51.68 $\pm$ 0.89} &
  \textbf{63.42 $\pm$ 1.02} \\
 &
  F1 &
  92.71 $\pm$ 0.44 &
  94.94 $\pm$ 0.36 &
  \textbf{67.66 $\pm$ 0.34} &
  91.57 $\pm$ 0.04 &
  \textbf{48.28 $\pm$ 0.26} &
  \textbf{54.28 $\pm$ 0.47} &
  \underline{54.14 $\pm$ 0.31} &
  \textbf{48.22 $\pm$ 2.17} &
  \textbf{63.22 $\pm$ 0.84} &
  \textbf{70.43 $\pm$ 0.61} &
  \textbf{51.43 $\pm$ 0.82} &
  \textbf{62.85 $\pm$ 1.06} \\ \hline
\end{tabular}%
}
\end{table*}

\subsubsection{Baselines}
We consider seven SOTA GNNs across both privacy-preserving and non-privacy-preserving settings.
\begin{itemize}
    \item \textit{Non-privacy-preserving GNNs}: We use a standard GCN \cite{kipf2016semi}, a representative GNN model without privacy protection.
    \item \textit{Link privacy-preserving GNNs}: We consider two SOTA models -- LapGraph \cite{wu2022linkteller} and PPRLG \cite{wang2021privacy}. LapGraph is based on differential privacy and PPRLG employs adversarial training.
    \item \textit{Node privacy-preserving GNNs}: This includes APGE \cite{li2020adversarial} and GAE \cite{kipf2016variational}. Both preserve privacy via adversarially optimizing sensitive attribute inference loss. 
    \item \textit{Dual privacy-preserving GNNs}: We consider DPGCN \cite{lin2022towards}, which employs local differential privacy, and GAE-NL, which adapts GAE \cite{kipf2016variational} by jointly optimizing graph reconstruction and sensitive attribute inference losses.
\end{itemize}

\subsubsection{Privacy Attack Setting}
We follow the attack setting in \cite{li2020adversarial}.
Specifically, the goal of the attacker is to infer users' sensitive attributes and link status given learned node embeddings and accessible adjacency matrix.

\noindent\textbf{Node-level Privacy Inference. }
Given a set of nodes $\mathcal{V}_a$ with known sensitive attribute $\mathrm{Y}_{s,a}$ and their learned node embeddings $\mathrm{Z}_a$, the attacker trains a surrogate model $f:\mathbb{R}^{M}\to\mathbb{R}^{C_s}$ to infer target users' sensitive attribute by minimizing the following loss:
\begin{equation}
    \min \sum_{v\in\mathcal{V}_a, y_v\in\mathrm{Y}_{s,a}} l(f(\mathrm{z}_v), y_v),
\end{equation}
where $l(\cdot)$ is a classification loss such as the cross entropy loss and $f$ is a multi-layer perceptron. With the trained model $f$, the attacker can then infer the  sensitive attributes of target nodes by exploiting their learned node embeddings.

\noindent\textbf{Link-level Privacy Inference. }
The attacker can obtain a set of positive links $\mathcal{E}_p\subset\mathcal{E}$ (where $\tilde{A}_{u,v}=1$) and a set of negative links $\mathcal{E}_n\not\subset\mathcal{E}$ (where $\tilde{A}_{u,v}=0$) as training links from an accessible adjacency matrix $\tilde{A}$. Based on this information, the attacker trains a link predictor $g:\mathbb{R}^{M}\times\mathbb{R}^{M}\to [0, 1]$ using training links and learned node embeddings $\mathrm{Z}$ with the following objective function:
\begin{equation}
    \min \sum_{(u, v)\in\mathcal{E}_p\cup\mathcal{E}_n} l(g(\mathrm{z}_u, \mathrm{z}_v),\tilde{A}_{u,v}),
\end{equation}
where $l(\cdot)$ is the cross entropy loss and $g$ is implemented as a multi-layer perceptron. With the trained model $g$, the attacker then infers the link status between unlabeled pairs of nodes. Note that the accessible matrix $\tilde{A}$ can differ from the real adjacency matrix $A$.

\subsubsection{Evaluation Metrics}
We consider evaluation metrics for both utility and privacy.
For utility, we use node classification accuracy and F1 score as the criteria. Higher accuracy and F1 score are desired in utility attribute prediction.
Evaluation of privacy preservation on graphs consists of both node and link levels. For node-level privacy, we use attacker’s sensitive attributes inference accuracy and F1 score as the metrics. For link-level privacy, we randomly sample 85\% positive links and negative links for training, 5\% for validating, and 10\% for testing, and measure attacker's link status inference accuracy and F1 score. For both tasks, lower accuracy and F1 score indicate higher privacy preservation.

\subsubsection{Parameter Setting}
The learning rates $\eta_1$ and $\eta_2$ are set to be $10^{-4}$ and 0.2, respectively. Following \cite{li2020adversarial}, we set both $\eta_3$ and $\eta_4$ to $10^{-2}$ for Yale4, Rochester38 and Recidivism datasets; for German dataset, $\eta_3$ and $\eta_4$ are set to $10^{-4}$. Hyperparameters $\alpha,\mu$, and $\lambda$ are set to 1 (To illustrate the parameter selection process, we give an exemplar analysis of $\alpha$ in Sec.~\ref{sec:ablation}). For each set of experiments, we report the average performance across 10 trials with standard deviations. Training epochs are set to 2000, 1500, 500 and 500 for the four datasets, respectively.
\subsection{Dual-Privacy Preservation in GNNs}
The first research question (\textbf{RQ. 1}) seeks to investigate the effectiveness of DPPGNN in terms of node and link privacy preservation. We compare all baselines on the four benchmark datasets. The results are in the last two columns of Table~\ref{tab:result}, with the best and second-best results highlighted in bold and underscored fonts, respectively.

We have the following three observations:
First, DPPGNN consistently achieves the best results w.r.t. link privacy preservation and outperforms all baselines in most datasets w.r.t. node privacy preservation. For example, it improves over the second best Acc. (DPGCN) in link-level privacy task by 10.8\% for the Rocherster38 dataset. LapGraph shows the second-best performance w.r.t. link-privacy preservation for the German Credit dataset. However, in the node classification task, its performance is greatly reduced while DPPGNN still shows competitive utility (69.83 vs. 73.32). Second, as expected, approaches that preserve either node or link privacy consistently show undesired privacy leakage w.r.t. the other task while DPPGNN can achieve better performance for both node and link privacy preservation. Take the node privacy-preserving GNN -- APGE -- as an example. It achieves a very competitive performance in protecting node sensitive information while showing undesired performance w.r.t. link privacy preservation, across all three datasets. Finally, most single privacy-preserving approaches do not outperform DPPGNN w.r.t. their target tasks. For example, DPPGNN outperforms the link privacy-preserving GNN -- PPRLG -- in link privacy preservation task.  
This suggests that node privacy and link privacy may be inherently interrelated, and it is critical to tackle both when developing privacy-preserving GNNs.

\subsection{Node Classification}
The other important evaluation examines the utility of privacy-preserving node embeddings, i.e., \textbf{RQ. 2}. The Utility column in Table~\ref{tab:result} presents the node classification results.

We have the following observations:
First, DPPGNN presents very competitive node classification performance for all three datasets. For instance, DPPGNN outperforms all baselines w.r.t. F1 score for the German Credit dataset. Second, compared to single privacy-preserving GNN baselines, DPPGNN achieves superior dual-privacy preservation performance while still showing similar classification performance. Take LapGraph as an example, when applied to the Rochester38 dataset, DPPGNN presents better performance for both node classification (93.89 vs. 90.11) and privacy preservation (node-level: 52.08 vs. 54.73; link-level: 63.78 vs. 80.63).
The two observations imply that the proposed model not only well preserves dual privacy but also achieves competitive model utility. Namely, our model achieves a better privacy-utility trade-off, thanks to the adversarial learning framework.

In summary, based on the results for \textbf{RQ. 1-2}, we validate the effectiveness of the proposed model in terms of dual-privacy preservation and utility maintenance in GNNs under structural bias. 

\begin{figure}[]
  \centering
  \includegraphics[width=\linewidth]{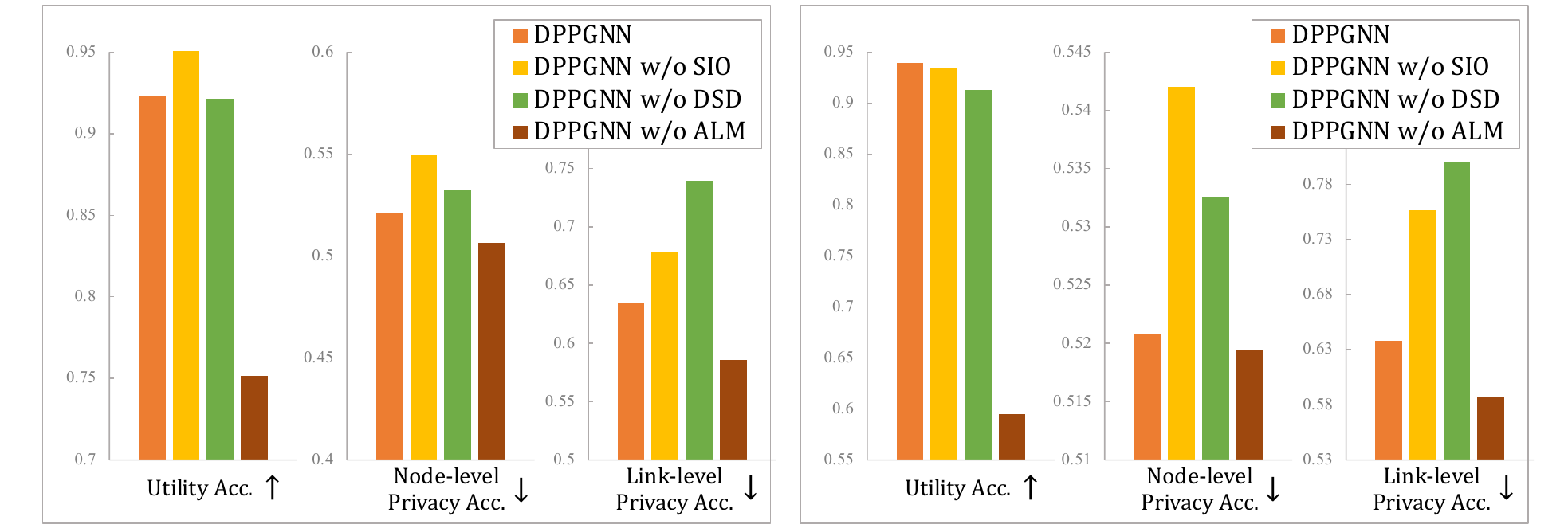}
  \vspace{-20pt}
  \caption{Ablation study on Recidivism (left) and Rochester38 (right).}
  \Description{ablation german.}
  \label{fig:ablation}
\end{figure}

\subsection{Ablation Study}\label{sec:ablation}

\begin{table}[]
\caption{Performance (accuracy) of node classification and dual-privacy preservation under different structural bias ($\alpha$).}\label{tab:SB-NL}
\centering
\vspace{-8pt}
\resizebox{\columnwidth}{!}{%
\begin{tabular}{c|ccc|ccc}
\hline
               & \multicolumn{3}{c|}{German Credit}                              & \multicolumn{3}{c}{Rochester38}                                 \\ \cline{2-7} 
 &
  \multicolumn{1}{l|}{\#utility} &
  \multicolumn{1}{c|}{\#node privacy} &
  \#link privacy &
  \multicolumn{1}{l|}{\#utility} &
  \multicolumn{1}{c|}{\#node privacy} &
  \#link privacy \\ \hline
$\alpha = 0.1$ & \multicolumn{1}{c|}{72.20} & \multicolumn{1}{c|}{69.80} & 51.67 & \multicolumn{1}{c|}{93.91} & \multicolumn{1}{c|}{52.53} & 64.20 \\
$\alpha = 1$   & \multicolumn{1}{c|}{71.80} & \multicolumn{1}{c|}{68.90} & 51.16 & \multicolumn{1}{c|}{94.32} & \multicolumn{1}{c|}{52.12} & 63.36 \\
$\alpha = 10$  & \multicolumn{1}{c|}{70.10} & \multicolumn{1}{c|}{68.70} & 50.95 & \multicolumn{1}{c|}{93.01} & \multicolumn{1}{c|}{51.93} & 61.12 \\ \hline
\end{tabular}%
}
\end{table}

\subsubsection{Effectiveness of Each Module}

To have a better understanding of the working mechanism of DPPGNN (i.e., \textbf{RQ. 3}), we further conduct ablation studies to investigate the impact of each module in the proposed framework on its performance. 
Specifically, we alternatively remove one of the three modules and compare its performance with the full DPPGNN. Results for the Recidivism dataset and Rochester38 dataset are shown in Fig.~\ref{fig:ablation}.

We can see that removing the SIO module improves utility but causes more privacy leakage on both sensitive attributes and link status.
This suggests that SIO can effectively obfuscate sensitive information in node representations, rendering stronger node- and link-level privacy preservation. Removing the DSD module has the largest influence on the link privacy preservation task, indicating the effectiveness of DSD for debiasing graph structure and dynamically purging private information from edges.  
Finally, we observe that ALM generally enables the framework to have a better privacy-utility trade-off, consistent with previous findings.  
These results demonstrate that each module is indispensable to the proposed framework to guarantee good performance in both dual-privacy preservation and downstream tasks.

\subsubsection{Impact of $\mathcal{L}_{DSD}$}
\begin{figure}[]
  \centering
  \includegraphics[width=\linewidth]{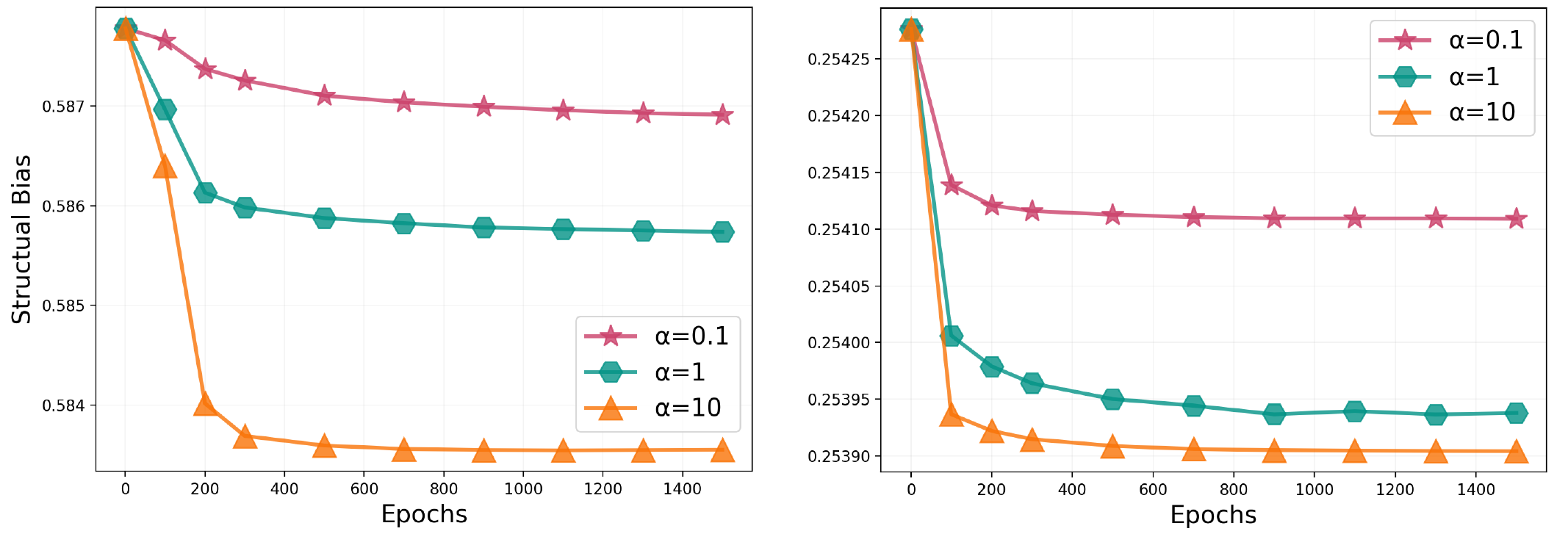}
  \vspace{-20pt}
  \caption{Change of Structural Bias during training process on German Credit (left) and Rochester38 (right).}
  \Description{structual bias.}
  \label{fig:sbv}
\end{figure}

We are also interested in understanding the structural debiasing function $\mathcal{L}_{DSD}$ in the DSD module. Particularly, we investigate (1) whether this dynamic debiasing approach can indeed reduce the structural bias, (2) how it affects DPPGNN's performance w.r.t. the node and link privacy preservation, and (3) the optimal $\alpha$ for overall performance. 
We vary $\alpha$, the parameter used to control the weight of $\mathcal{L}_{adj}^{(s)}$ in Eq. \ref{obj:dsd}, among $\{0.1, 1, 10\}$, and present the results for German Credit and Rochester38 datasets. As shown in Fig. \ref{fig:sbv}, the structural bias in graphs decreases as the training progress. This applies to all different values of $\alpha$, suggesting that $\mathcal{L}_{adj}^{(s)}$ is effective to reduce structural bias. Results for dual privacy preservation can be seen in Table~\ref{tab:SB-NL}. Of particular interest is that when $\alpha$ increases, i.e., less biased structure, both node and link privacy leakage decrease. This suggests that message passing under structural bias can indeed lead to more severe privacy leakage at both node and link levels, and DSD can effectively debias graph structure. For utility comparison, we see that $\alpha=1$ can in general maintains competitive utility for both datasets. We, therefore, set $\alpha=1$ in DPPGNN to get a better utility-privacy trade-off.

\section{Related Work}
Privacy is one of the pillars of socially responsible AI \cite{cheng2021socially,cheng2023socially}. Here, we review it from both attacks and preservation perspectives.\\
\textbf{Privacy Attacks on GNNs.}
With GNNs' wide applications in various domains, many privacy attack techniques for GNNs have been proposed to investigate their privacy issue. There are generally three kinds of attacks depending on the attacker's goal: membership inference attack \cite{shokri2017membership,salem2018ml}, model extraction attack \cite{tramer2016stealing,hu2022tp,juuti2019prada}, 
and attribute inference attack \cite{duddu2020quantifying,gong2016you,he2020stealing,zhang2021graphmi,zhang2022inference}. 
Membership inference attacks aim to identify if a node is in the training dataset, which can compromise the node’s privacy by revealing its sensitive information. For example, if the adversary learns that a patient’s clinical record was part of the training dataset for a pneumonia treatment model, s/he can deduce that the patient has pneumonia \cite{shokri2017membership}. Model extraction attacks seek to extract information on target model's parameters and reconstruct one substitute model that mimics its behavior.
Attribute inference attacks aim to expose the private information of a graph used to train a model. The attack can target either the link status or the node attributes of the graph. The former aims to detect the presence of an edge between any two nodes, while the latter tries to infer the sensitive attributes of a node, such as nationality or political view.
Overall, these works underline various privacy risks associated with GNNs and demonstrate the vulnerability of GNN models to privacy attacks.

\noindent\textbf{Privacy-Preserving GNNs.}
Correspondingly, privacy-preserving GNNs have been proposed in response to the attack techniques. One line of research is \textit{encryption} \cite{fang2021privacy}. However, encryption brings an extra cost of time and computation. The second direction is \textit{adversarial training}, which maximizes model utility and minimizes privacy loss simultaneously in the training process. For example, Liao et al. \cite{liao2020graph} proposed an adversarial learning approach where they introduced a minimax game between the desired graph feature encoder and the worst-case attacker. Li et al. \cite{li2020adversarial} presented a
graph adversarial training framework that integrated disentangling
and purging mechanisms to remove users’ private information from
learned node representations. However, both of these works only consider node privacy. Another research direction is \textit{federated and split learning}. In \cite{zhou2020privacy}, for example, the authors
studied privacy-preserving node classification by splitting the computation graph of a GNN among multiple data holders. They further used a trusted server to combine the information from different parties and complete the training process. Mei et al. \cite{mei2019sgnn} proposed to use structural similarity and federated learning to hide users' private information in GNNs. Federated GNN is also used in privacy-preserving recommendations \cite{wu2021fedgnn}. However, these approaches rely on the existence of a trusted server for model aggregation. In practice, it is difficult to find a trusted party to store users' sensitive information. The fourth direction leverages \textit{differential privacy} to protect privacy in GNNs. For instance, Sina et al. \cite{sajadmanesh2020differential} introduced node privacy in GNNs through local differential privacy \cite{kasiviswanathan2011can}. Xu et al. \cite{xu2018dpne} proposed private graph embedding via perturbing the loss function. However, compared to adversarial learning, differential privacy might lead to greater utility loss due to added noise.

To summarize, the existing research on privacy-preserving GNNs primarily concentrates on safeguarding privacy at the node and/or link level, neglecting the underlying mechanisms contributing to such privacy leakage. Our study aims to bridge this gap by offering the first theoretical analysis and simulation studies that illustrate how GNNs can compromise privacy by biased graph structures during the message-passing process. To tackle this issue, we propose a solution that ensures the preservation of both node and link privacy by obscuring sensitive information and mitigating the bias within the dynamic structure under an adversarial learning framework. The proposed model effectively balances the trade-off between maintaining privacy and preserving the utility of the model.


\section{Conclusion}
This work investigates the crucial role of message passing in GNNs in both node and link privacy leakage on graphs. We provide the first thorough theoretical examinations and simulation studies to demonstrate that the primary message-passing process in GNNs can spread and amplify privacy breaches on graphs, and such breaches exacerbate when the graph becomes more structurally biased. To address this issue, we introduce a principled dual-privacy-preserving GNN framework -- DPPGNN. It uses sensitive information obfuscation and dynamic structure debiasing to remove sensitive information from both nodes and links, and achieves better privacy-utility trade-off through adversarial training. The experimental results show that DPPGNN excels in preserving dual privacy while still delivering competitive utility in downstream tasks.
\section*{Acknowledgements}
This material is in part supported by the Cisco Research Gift Grant.
\appendix
\section{Proof of Proposition~\ref{pp1}}
\label{appdix:1}

\noindent(1) 
\textbf{When does privacy leakage become more severe after GCN-like message passing?}

If privacy leakage becomes more severe after message passing, we shall have $\Delta PL = PL'-PL>0$, i.e.,
\begin{equation}\label{eq:26}
    \Delta PL = \frac{k\mu^2}{2} \cdot\left ( \frac{n(p+q)(np-nq+1)^2}{1+n(p+q)} - 1  \right ) > 0,
\end{equation}

Recall the definition of structural bias in Def. \ref{def:sb}, for synthetic graph $\mathcal{G}$, $B=\frac{p-q}{p+q}$. Then, inequality~(\ref{eq:26}) is equivalent to:
\begin{equation}
\label{eq:add}
    n^3(p+q)^3B^2+2n^2(p+q)^2B-1>0.
\end{equation}

The above inequality holds when:
\begin{equation}
    B>\frac{\sqrt{n^2(p+q)^2+n(p+q)}-n(p+q)}{n^2(p+q)^2}.
\end{equation}

That is to say, privacy leakage becomes more severe when structural bias $B>\frac{\sqrt{n^2(p+q)^2+n(p+q)}-n(p+q)}{n^2(p+q)^2}$.

\noindent(2)
\textbf{ How do different structural biases affect privacy leakage after GCN-like message passing?}

The difference in privacy leakage before and after message passing could be written as:
\begin{equation}
    \Delta PL = \frac{k\mu^2}{2}\cdot \left ( \frac{n(p+q)\cdot \left(n(p+q)\cdot B+1\right)^2-n(p+q)-1}{1+n(p+q)} \right ).
\end{equation}

Since $n(p+q)$ is larger than zero identically, with a fixed $\mu$ and $k$, if $B_1>B_2$, then $\Delta PL_1>\Delta PL_2$.
That is, a larger structural bias leads to more privacy leakage after message passing on graphs.

\bibliographystyle{ACM-Reference-Format}
\balance
\bibliography{Reference}


\begin{thebibliography}{52}


\ifx \showCODEN    \undefined \def \showCODEN     #1{\unskip}     \fi
\ifx \showDOI      \undefined \def \showDOI       #1{#1}\fi
\ifx \showISBNx    \undefined \def \showISBNx     #1{\unskip}     \fi
\ifx \showISBNxiii \undefined \def \showISBNxiii  #1{\unskip}     \fi
\ifx \showISSN     \undefined \def \showISSN      #1{\unskip}     \fi
\ifx \showLCCN     \undefined \def \showLCCN      #1{\unskip}     \fi
\ifx \shownote     \undefined \def \shownote      #1{#1}          \fi
\ifx \showarticletitle \undefined \def \showarticletitle #1{#1}   \fi
\ifx \showURL      \undefined \def \showURL       {\relax}        \fi
\providecommand\bibfield[2]{#2}
\providecommand\bibinfo[2]{#2}
\providecommand\natexlab[1]{#1}
\providecommand\showeprint[2][]{arXiv:#2}

\bibitem[Agarwal et~al\mbox{.}(2021)]%
        {agarwal2021towards}
\bibfield{author}{\bibinfo{person}{Chirag Agarwal}, \bibinfo{person}{Himabindu
  Lakkaraju}, {and} \bibinfo{person}{Marinka Zitnik}.}
  \bibinfo{year}{2021}\natexlab{}.
\newblock \showarticletitle{Towards a Unified Framework for Fair and Stable
  Graph Representation Learning}.
\newblock \bibinfo{journal}{\emph{arXiv preprint arXiv:2102.13186}}
  (\bibinfo{year}{2021}).
\newblock


\bibitem[Arjovsky et~al\mbox{.}(2017)]%
        {arjovsky2017wasserstein}
\bibfield{author}{\bibinfo{person}{Martin Arjovsky}, \bibinfo{person}{Soumith
  Chintala}, {and} \bibinfo{person}{L{\'e}on Bottou}.}
  \bibinfo{year}{2017}\natexlab{}.
\newblock \showarticletitle{Wasserstein generative adversarial networks}. In
  \bibinfo{booktitle}{\emph{International conference on machine learning}}.
  PMLR, \bibinfo{pages}{214--223}.
\newblock


\bibitem[Bhattacharyya(1943)]%
        {bhattacharyya1943measure}
\bibfield{author}{\bibinfo{person}{Anil Bhattacharyya}.}
  \bibinfo{year}{1943}\natexlab{}.
\newblock \showarticletitle{On a measure of divergence between two statistical
  populations defined by their probability distribution}.
\newblock \bibinfo{journal}{\emph{Bulletin of the Calcutta Mathematical
  Society}}  \bibinfo{volume}{35} (\bibinfo{year}{1943}),
  \bibinfo{pages}{99--110}.
\newblock


\bibitem[Bhattacharyya(1946)]%
        {bhattacharyya1946measure}
\bibfield{author}{\bibinfo{person}{Anil Bhattacharyya}.}
  \bibinfo{year}{1946}\natexlab{}.
\newblock \showarticletitle{On a measure of divergence between two multinomial
  populations}.
\newblock \bibinfo{journal}{\emph{Sankhy{\=a}: the indian journal of
  statistics}} (\bibinfo{year}{1946}), \bibinfo{pages}{401--406}.
\newblock


\bibitem[Cheng et~al\mbox{.}(2022)]%
        {cheng2022financial}
\bibfield{author}{\bibinfo{person}{Dawei Cheng}, \bibinfo{person}{Fangzhou
  Yang}, \bibinfo{person}{Sheng Xiang}, {and} \bibinfo{person}{Jin Liu}.}
  \bibinfo{year}{2022}\natexlab{}.
\newblock \showarticletitle{Financial time series forecasting with
  multi-modality graph neural network}.
\newblock \bibinfo{journal}{\emph{Pattern Recognition}}  \bibinfo{volume}{121}
  (\bibinfo{year}{2022}), \bibinfo{pages}{108218}.
\newblock


\bibitem[Cheng and Liu(2023)]%
        {cheng2023socially}
\bibfield{author}{\bibinfo{person}{Lu Cheng} {and} \bibinfo{person}{Huan Liu}.}
  \bibinfo{year}{2023}\natexlab{}.
\newblock \bibinfo{booktitle}{\emph{Socially Responsible AI: Theories and
  Practices}}.
\newblock \bibinfo{publisher}{World Scientific}.
\newblock


\bibitem[Cheng et~al\mbox{.}(2021)]%
        {cheng2021socially}
\bibfield{author}{\bibinfo{person}{Lu Cheng}, \bibinfo{person}{Kush~R
  Varshney}, {and} \bibinfo{person}{Huan Liu}.}
  \bibinfo{year}{2021}\natexlab{}.
\newblock \showarticletitle{Socially responsible ai algorithms: Issues,
  purposes, and challenges}.
\newblock \bibinfo{journal}{\emph{Journal of Artificial Intelligence Research}}
   \bibinfo{volume}{71} (\bibinfo{year}{2021}), \bibinfo{pages}{1137--1181}.
\newblock


\bibitem[Deshpande et~al\mbox{.}(2018)]%
        {deshpande2018contextual}
\bibfield{author}{\bibinfo{person}{Yash Deshpande}, \bibinfo{person}{Subhabrata
  Sen}, \bibinfo{person}{Andrea Montanari}, {and} \bibinfo{person}{Elchanan
  Mossel}.} \bibinfo{year}{2018}\natexlab{}.
\newblock \showarticletitle{Contextual stochastic block models}.
\newblock \bibinfo{journal}{\emph{Advances in Neural Information Processing
  Systems}}  \bibinfo{volume}{31} (\bibinfo{year}{2018}).
\newblock


\bibitem[Diehl et~al\mbox{.}(2019)]%
        {diehl2019graph}
\bibfield{author}{\bibinfo{person}{Frederik Diehl}, \bibinfo{person}{Thomas
  Brunner}, \bibinfo{person}{Michael~Truong Le}, {and} \bibinfo{person}{Alois
  Knoll}.} \bibinfo{year}{2019}\natexlab{}.
\newblock \showarticletitle{Graph neural networks for modelling traffic
  participant interaction}. In \bibinfo{booktitle}{\emph{2019 IEEE Intelligent
  Vehicles Symposium (IV)}}. IEEE, \bibinfo{pages}{695--701}.
\newblock


\bibitem[Duddu et~al\mbox{.}(2020)]%
        {duddu2020quantifying}
\bibfield{author}{\bibinfo{person}{Vasisht Duddu}, \bibinfo{person}{Antoine
  Boutet}, {and} \bibinfo{person}{Virat Shejwalkar}.}
  \bibinfo{year}{2020}\natexlab{}.
\newblock \showarticletitle{Quantifying Privacy Leakage in Graph Embedding}.
\newblock \bibinfo{journal}{\emph{arXiv preprint arXiv:2010.00906}}
  (\bibinfo{year}{2020}).
\newblock


\bibitem[Fang and Qian(2021)]%
        {fang2021privacy}
\bibfield{author}{\bibinfo{person}{Haokun Fang} {and} \bibinfo{person}{Quan
  Qian}.} \bibinfo{year}{2021}\natexlab{}.
\newblock \showarticletitle{Privacy Preserving Machine Learning with
  Homomorphic Encryption and Federated Learning}.
\newblock \bibinfo{journal}{\emph{Future Internet}} \bibinfo{volume}{13},
  \bibinfo{number}{4} (\bibinfo{year}{2021}), \bibinfo{pages}{94}.
\newblock


\bibitem[Gaudelet et~al\mbox{.}(2021)]%
        {gaudelet2021utilizing}
\bibfield{author}{\bibinfo{person}{Thomas Gaudelet}, \bibinfo{person}{Ben Day},
  \bibinfo{person}{Arian~R Jamasb}, \bibinfo{person}{Jyothish Soman},
  \bibinfo{person}{Cristian Regep}, \bibinfo{person}{Gertrude Liu},
  \bibinfo{person}{Jeremy~BR Hayter}, \bibinfo{person}{Richard Vickers},
  \bibinfo{person}{Charles Roberts}, \bibinfo{person}{Jian Tang},
  {et~al\mbox{.}}} \bibinfo{year}{2021}\natexlab{}.
\newblock \showarticletitle{Utilizing graph machine learning within drug
  discovery and development}.
\newblock \bibinfo{journal}{\emph{Briefings in bioinformatics}}
  \bibinfo{volume}{22}, \bibinfo{number}{6} (\bibinfo{year}{2021}),
  \bibinfo{pages}{bbab159}.
\newblock


\bibitem[Gilmer et~al\mbox{.}(2017)]%
        {gilmer2017neural}
\bibfield{author}{\bibinfo{person}{Justin Gilmer}, \bibinfo{person}{Samuel~S
  Schoenholz}, \bibinfo{person}{Patrick~F Riley}, \bibinfo{person}{Oriol
  Vinyals}, {and} \bibinfo{person}{George~E Dahl}.}
  \bibinfo{year}{2017}\natexlab{}.
\newblock \showarticletitle{Neural message passing for quantum chemistry}. In
  \bibinfo{booktitle}{\emph{International conference on machine learning}}.
  PMLR, \bibinfo{pages}{1263--1272}.
\newblock


\bibitem[Gong and Liu(2016)]%
        {gong2016you}
\bibfield{author}{\bibinfo{person}{Neil~Zhenqiang Gong} {and}
  \bibinfo{person}{Bin Liu}.} \bibinfo{year}{2016}\natexlab{}.
\newblock \showarticletitle{You are who you know and how you behave: Attribute
  inference attacks via users' social friends and behaviors}. In
  \bibinfo{booktitle}{\emph{25th $\{$USENIX$\}$ Security Symposium
  ($\{$USENIX$\}$ Security 16)}}. \bibinfo{pages}{979--995}.
\newblock


\bibitem[He et~al\mbox{.}(2020)]%
        {he2020stealing}
\bibfield{author}{\bibinfo{person}{Xinlei He}, \bibinfo{person}{Jinyuan Jia},
  \bibinfo{person}{Michael Backes}, \bibinfo{person}{Neil~Zhenqiang Gong},
  {and} \bibinfo{person}{Yang Zhang}.} \bibinfo{year}{2020}\natexlab{}.
\newblock \showarticletitle{Stealing Links from Graph Neural Networks}.
\newblock \bibinfo{journal}{\emph{arXiv preprint arXiv:2005.02131}}
  (\bibinfo{year}{2020}).
\newblock


\bibitem[He et~al\mbox{.}(2021)]%
        {he2021node}
\bibfield{author}{\bibinfo{person}{Xinlei He}, \bibinfo{person}{Rui Wen}, {and}
  \bibinfo{person}{et~al. Wu}.} \bibinfo{year}{2021}\natexlab{}.
\newblock \showarticletitle{Node-Level Membership Inference Attacks Against
  Graph Neural Networks}.
\newblock \bibinfo{journal}{\emph{arXiv preprint arXiv:2102.05429}}
  (\bibinfo{year}{2021}).
\newblock


\bibitem[Hu et~al\mbox{.}(2022a)]%
        {hu2022learning}
\bibfield{author}{\bibinfo{person}{Hui Hu}, \bibinfo{person}{Lu Cheng},
  \bibinfo{person}{Jayden~Parker Vap}, {and} \bibinfo{person}{Mike Borowczak}.}
  \bibinfo{year}{2022}\natexlab{a}.
\newblock \showarticletitle{Learning Privacy-Preserving Graph Convolutional
  Network with Partially Observed Sensitive Attributes}. In
  \bibinfo{booktitle}{\emph{Proceedings of the ACM Web Conference 2022}}.
  \bibinfo{pages}{3552--3561}.
\newblock


\bibitem[Hu et~al\mbox{.}(2022b)]%
        {hu2022tp}
\bibfield{author}{\bibinfo{person}{Hui Hu}, \bibinfo{person}{Jessa
  Gegax-Randazzo}, \bibinfo{person}{Clay Carper}, {and} \bibinfo{person}{Mike
  Borowczak}.} \bibinfo{year}{2022}\natexlab{b}.
\newblock \showarticletitle{TP-NET: Training Privacy-Preserving Deep Neural
  Networks under Side-Channel Power Attacks}. In \bibinfo{booktitle}{\emph{2022
  IEEE International Symposium on Smart Electronic Systems (iSES)}}. IEEE,
  \bibinfo{pages}{439--444}.
\newblock


\bibitem[Janson et~al\mbox{.}(2011)]%
        {janson2011random}
\bibfield{author}{\bibinfo{person}{Svante Janson}, \bibinfo{person}{Andrzej
  Rucinski}, {and} \bibinfo{person}{Tomasz Luczak}.}
  \bibinfo{year}{2011}\natexlab{}.
\newblock \bibinfo{booktitle}{\emph{Random graphs}}.
\newblock \bibinfo{publisher}{John Wiley \& Sons}.
\newblock


\bibitem[Juuti et~al\mbox{.}(2019)]%
        {juuti2019prada}
\bibfield{author}{\bibinfo{person}{Mika Juuti}, \bibinfo{person}{Sebastian
  Szyller}, \bibinfo{person}{Samuel Marchal}, {and} \bibinfo{person}{N
  Asokan}.} \bibinfo{year}{2019}\natexlab{}.
\newblock \showarticletitle{PRADA: protecting against DNN model stealing
  attacks}. In \bibinfo{booktitle}{\emph{2019 IEEE European Symposium on
  Security and Privacy (EuroS\&P)}}. IEEE, \bibinfo{pages}{512--527}.
\newblock


\bibitem[Kasiviswanathan et~al\mbox{.}(2011)]%
        {kasiviswanathan2011can}
\bibfield{author}{\bibinfo{person}{Shiva~Prasad Kasiviswanathan},
  \bibinfo{person}{Homin~K Lee}, \bibinfo{person}{Kobbi Nissim},
  \bibinfo{person}{Sofya Raskhodnikova}, {and} \bibinfo{person}{Adam Smith}.}
  \bibinfo{year}{2011}\natexlab{}.
\newblock \showarticletitle{What can we learn privately?}
\newblock \bibinfo{journal}{\emph{SIAM J. Comput.}} \bibinfo{volume}{40},
  \bibinfo{number}{3} (\bibinfo{year}{2011}), \bibinfo{pages}{793--826}.
\newblock


\bibitem[Khanam et~al\mbox{.}(2022)]%
        {khanam2022homophily}
\bibfield{author}{\bibinfo{person}{Kazi~Zainab Khanam}, \bibinfo{person}{Gautam
  Srivastava}, {and} \bibinfo{person}{Vijay Mago}.}
  \bibinfo{year}{2022}\natexlab{}.
\newblock \showarticletitle{The homophily principle in social network analysis:
  A survey}.
\newblock \bibinfo{journal}{\emph{Multimedia Tools and Applications}}
  (\bibinfo{year}{2022}), \bibinfo{pages}{1--44}.
\newblock


\bibitem[Kipf and Welling(2016a)]%
        {kipf2016semi}
\bibfield{author}{\bibinfo{person}{Thomas~N Kipf} {and} \bibinfo{person}{Max
  Welling}.} \bibinfo{year}{2016}\natexlab{a}.
\newblock \showarticletitle{Semi-supervised classification with graph
  convolutional networks}.
\newblock \bibinfo{journal}{\emph{arXiv preprint arXiv:1609.02907}}
  (\bibinfo{year}{2016}).
\newblock


\bibitem[Kipf and Welling(2016b)]%
        {kipf2016variational}
\bibfield{author}{\bibinfo{person}{Thomas~N Kipf} {and} \bibinfo{person}{Max
  Welling}.} \bibinfo{year}{2016}\natexlab{b}.
\newblock \showarticletitle{Variational graph auto-encoders}.
\newblock \bibinfo{journal}{\emph{arXiv preprint arXiv:1611.07308}}
  (\bibinfo{year}{2016}).
\newblock


\bibitem[Li et~al\mbox{.}(2020)]%
        {li2020adversarial}
\bibfield{author}{\bibinfo{person}{Kaiyang Li}, \bibinfo{person}{Guangchun
  Luo}, \bibinfo{person}{Yang Ye}, \bibinfo{person}{Wei Li},
  \bibinfo{person}{Shihao Ji}, {and} \bibinfo{person}{Zhipeng Cai}.}
  \bibinfo{year}{2020}\natexlab{}.
\newblock \showarticletitle{Adversarial Privacy Preserving Graph Embedding
  against Inference Attack}.
\newblock \bibinfo{journal}{\emph{IEEE Internet of Things Journal}}
  (\bibinfo{year}{2020}).
\newblock


\bibitem[Li et~al\mbox{.}(2021)]%
        {li2021dyadic}
\bibfield{author}{\bibinfo{person}{Peizhao Li}, \bibinfo{person}{Yifei Wang},
  \bibinfo{person}{Han Zhao}, \bibinfo{person}{Pengyu Hong}, {and}
  \bibinfo{person}{Hongfu Liu}.} \bibinfo{year}{2021}\natexlab{}.
\newblock \showarticletitle{On dyadic fairness: Exploring and mitigating bias
  in graph connections}. In \bibinfo{booktitle}{\emph{International Conference
  on Learning Representations}}.
\newblock


\bibitem[Liao et~al\mbox{.}(2020)]%
        {liao2020graph}
\bibfield{author}{\bibinfo{person}{Peiyuan Liao}, \bibinfo{person}{Han Zhao},
  \bibinfo{person}{Keyulu Xu}, \bibinfo{person}{Tommi Jaakkola},
  \bibinfo{person}{Geoffrey Gordon}, \bibinfo{person}{Stefanie Jegelka}, {and}
  \bibinfo{person}{Ruslan Salakhutdinov}.} \bibinfo{year}{2020}\natexlab{}.
\newblock \showarticletitle{Graph Adversarial Networks: Protecting Information
  against Adversarial Attacks}.
\newblock \bibinfo{journal}{\emph{arXiv preprint arXiv:2009.13504}}
  (\bibinfo{year}{2020}).
\newblock


\bibitem[Lin et~al\mbox{.}(2022)]%
        {lin2022towards}
\bibfield{author}{\bibinfo{person}{Wanyu Lin}, \bibinfo{person}{Baochun Li},
  {and} \bibinfo{person}{Cong Wang}.} \bibinfo{year}{2022}\natexlab{}.
\newblock \showarticletitle{Towards private learning on decentralized graphs
  with local differential privacy}.
\newblock \bibinfo{journal}{\emph{arXiv preprint arXiv:2201.09398}}
  (\bibinfo{year}{2022}).
\newblock


\bibitem[Mei et~al\mbox{.}(2019)]%
        {mei2019sgnn}
\bibfield{author}{\bibinfo{person}{Guangxu Mei}, \bibinfo{person}{Ziyu Guo},
  \bibinfo{person}{Shijun Liu}, {and} \bibinfo{person}{Li Pan}.}
  \bibinfo{year}{2019}\natexlab{}.
\newblock \showarticletitle{Sgnn: A graph neural network based federated
  learning approach by hiding structure}. In \bibinfo{booktitle}{\emph{Big
  Data}}. IEEE, \bibinfo{pages}{2560--2568}.
\newblock


\bibitem[Merton(1988)]%
        {merton1988matthew}
\bibfield{author}{\bibinfo{person}{Robert~K Merton}.}
  \bibinfo{year}{1988}\natexlab{}.
\newblock \showarticletitle{The Matthew effect in science, II: Cumulative
  advantage and the symbolism of intellectual property}.
\newblock \bibinfo{journal}{\emph{isis}} \bibinfo{volume}{79},
  \bibinfo{number}{4} (\bibinfo{year}{1988}), \bibinfo{pages}{606--623}.
\newblock


\bibitem[Olatunji et~al\mbox{.}(2021)]%
        {olatunji2021membership}
\bibfield{author}{\bibinfo{person}{Iyiola~E Olatunji},
  \bibinfo{person}{Wolfgang Nejdl}, {and} \bibinfo{person}{Megha Khosla}.}
  \bibinfo{year}{2021}\natexlab{}.
\newblock \showarticletitle{Membership inference attack on graph neural
  networks}. In \bibinfo{booktitle}{\emph{2021 Third IEEE International
  Conference on Trust, Privacy and Security in Intelligent Systems and
  Applications (TPS-ISA)}}. IEEE, \bibinfo{pages}{11--20}.
\newblock


\bibitem[Sajadmanesh and Gatica-Perez(2020)]%
        {sajadmanesh2020differential}
\bibfield{author}{\bibinfo{person}{Sina Sajadmanesh} {and}
  \bibinfo{person}{Daniel Gatica-Perez}.} \bibinfo{year}{2020}\natexlab{}.
\newblock \showarticletitle{When Differential Privacy Meets Graph Neural
  Networks}.
\newblock \bibinfo{journal}{\emph{arXiv preprint arXiv:2006.05535}}
  (\bibinfo{year}{2020}).
\newblock


\bibitem[Salem et~al\mbox{.}(2018)]%
        {salem2018ml}
\bibfield{author}{\bibinfo{person}{Ahmed Salem}, \bibinfo{person}{Yang Zhang},
  \bibinfo{person}{Mathias Humbert}, \bibinfo{person}{Pascal Berrang},
  \bibinfo{person}{Mario Fritz}, {and} \bibinfo{person}{Michael Backes}.}
  \bibinfo{year}{2018}\natexlab{}.
\newblock \showarticletitle{Ml-leaks: Model and data independent membership
  inference attacks and defenses on machine learning models}.
\newblock \bibinfo{journal}{\emph{arXiv preprint arXiv:1806.01246}}
  (\bibinfo{year}{2018}).
\newblock


\bibitem[Shokri et~al\mbox{.}(2017)]%
        {shokri2017membership}
\bibfield{author}{\bibinfo{person}{Reza Shokri}, \bibinfo{person}{Marco
  Stronati}, \bibinfo{person}{Congzheng Song}, {and} \bibinfo{person}{Vitaly
  Shmatikov}.} \bibinfo{year}{2017}\natexlab{}.
\newblock \showarticletitle{Membership inference attacks against machine
  learning models}. In \bibinfo{booktitle}{\emph{2017 IEEE Symposium on
  Security and Privacy (SP)}}. IEEE, \bibinfo{pages}{3--18}.
\newblock


\bibitem[Tram{\`e}r et~al\mbox{.}(2016)]%
        {tramer2016stealing}
\bibfield{author}{\bibinfo{person}{Florian Tram{\`e}r}, \bibinfo{person}{Fan
  Zhang}, \bibinfo{person}{Ari Juels}, \bibinfo{person}{Michael~K Reiter},
  {and} \bibinfo{person}{Thomas Ristenpart}.} \bibinfo{year}{2016}\natexlab{}.
\newblock \showarticletitle{Stealing machine learning models via prediction
  apis}. In \bibinfo{booktitle}{\emph{25th $\{$USENIX$\}$ Security Symposium
  ($\{$USENIX$\}$ Security 16)}}. \bibinfo{pages}{601--618}.
\newblock


\bibitem[Traud et~al\mbox{.}(2012)]%
        {traud2012social}
\bibfield{author}{\bibinfo{person}{Amanda~L Traud}, \bibinfo{person}{Peter~J
  Mucha}, {and} \bibinfo{person}{Mason~A Porter}.}
  \bibinfo{year}{2012}\natexlab{}.
\newblock \showarticletitle{Social structure of facebook networks}.
\newblock \bibinfo{journal}{\emph{Physica A: Statistical Mechanics and its
  Applications}} \bibinfo{volume}{391}, \bibinfo{number}{16}
  (\bibinfo{year}{2012}), \bibinfo{pages}{4165--4180}.
\newblock


\bibitem[Tripathy et~al\mbox{.}(2019)]%
        {tripathy2019privacy}
\bibfield{author}{\bibinfo{person}{Ardhendu Tripathy}, \bibinfo{person}{Ye
  Wang}, {and} \bibinfo{person}{Prakash Ishwar}.}
  \bibinfo{year}{2019}\natexlab{}.
\newblock \showarticletitle{Privacy-preserving adversarial networks}. In
  \bibinfo{booktitle}{\emph{2019 57th Annual Allerton Conference on
  Communication, Control, and Computing (Allerton)}}. IEEE,
  \bibinfo{pages}{495--505}.
\newblock


\bibitem[Villani(2009)]%
        {villani2009optimal}
\bibfield{author}{\bibinfo{person}{C{\'e}dric Villani}.}
  \bibinfo{year}{2009}\natexlab{}.
\newblock \bibinfo{booktitle}{\emph{Optimal transport: old and new}}.
  Vol.~\bibinfo{volume}{338}.
\newblock \bibinfo{publisher}{Springer}.
\newblock


\bibitem[Wang et~al\mbox{.}(2021)]%
        {wang2021privacy}
\bibfield{author}{\bibinfo{person}{Binghui Wang}, \bibinfo{person}{Jiayi Guo},
  \bibinfo{person}{Ang Li}, \bibinfo{person}{Yiran Chen}, {and}
  \bibinfo{person}{Hai Li}.} \bibinfo{year}{2021}\natexlab{}.
\newblock \showarticletitle{Privacy-Preserving Representation Learning on
  Graphs: A Mutual Information Perspective}.
\newblock \bibinfo{journal}{\emph{arXiv preprint arXiv:2107.01475}}
  (\bibinfo{year}{2021}).
\newblock


\bibitem[Wang et~al\mbox{.}(2023)]%
        {wang2023causalse}
\bibfield{author}{\bibinfo{person}{Qianru Wang}, \bibinfo{person}{Bin Guo},
  \bibinfo{person}{Lu Cheng}, \bibinfo{person}{Zhiwen Yu}, {and}
  \bibinfo{person}{Huan Liu}.} \bibinfo{year}{2023}\natexlab{}.
\newblock \showarticletitle{CausalSE: Understanding Varied Spatial Effects with
  Missing Data Toward Adding New Bike-sharing Stations}.
\newblock \bibinfo{journal}{\emph{ACM Transactions on Knowledge Discovery from
  Data}} \bibinfo{volume}{17}, \bibinfo{number}{2} (\bibinfo{year}{2023}),
  \bibinfo{pages}{1--24}.
\newblock


\bibitem[Wang and Carreira-Perpin{\'a}n(2013)]%
        {wang2013projection}
\bibfield{author}{\bibinfo{person}{Weiran Wang} {and} \bibinfo{person}{Miguel~A
  Carreira-Perpin{\'a}n}.} \bibinfo{year}{2013}\natexlab{}.
\newblock \showarticletitle{Projection onto the probability simplex: An
  efficient algorithm with a simple proof, and an application}.
\newblock \bibinfo{journal}{\emph{arXiv preprint arXiv:1309.1541}}
  (\bibinfo{year}{2013}).
\newblock


\bibitem[Wang et~al\mbox{.}(2020)]%
        {wang2020traffic}
\bibfield{author}{\bibinfo{person}{Xiaoyang Wang}, \bibinfo{person}{Yao Ma},
  \bibinfo{person}{Yiqi Wang}, \bibinfo{person}{Wei Jin}, \bibinfo{person}{Xin
  Wang}, \bibinfo{person}{Jiliang Tang}, \bibinfo{person}{Caiyan Jia}, {and}
  \bibinfo{person}{Jian Yu}.} \bibinfo{year}{2020}\natexlab{}.
\newblock \showarticletitle{Traffic flow prediction via spatial temporal graph
  neural network}. In \bibinfo{booktitle}{\emph{Proceedings of the web
  conference 2020}}. \bibinfo{pages}{1082--1092}.
\newblock


\bibitem[Wu et~al\mbox{.}(2021b)]%
        {wu2021adapting}
\bibfield{author}{\bibinfo{person}{Bang Wu}, \bibinfo{person}{Xiangwen Yang},
  \bibinfo{person}{Shirui Pan}, {and} \bibinfo{person}{Xingliang Yuan}.}
  \bibinfo{year}{2021}\natexlab{b}.
\newblock \showarticletitle{Adapting membership inference attacks to gnn for
  graph classification: Approaches and implications}. In
  \bibinfo{booktitle}{\emph{2021 IEEE International Conference on Data Mining
  (ICDM)}}. IEEE, \bibinfo{pages}{1421--1426}.
\newblock


\bibitem[Wu et~al\mbox{.}(2021a)]%
        {wu2021fedgnn}
\bibfield{author}{\bibinfo{person}{Chuhan Wu}, \bibinfo{person}{Fangzhao Wu},
  \bibinfo{person}{Yang Cao}, \bibinfo{person}{Yongfeng Huang}, {and}
  \bibinfo{person}{Xing Xie}.} \bibinfo{year}{2021}\natexlab{a}.
\newblock \showarticletitle{Fedgnn: Federated graph neural network for
  privacy-preserving recommendation}.
\newblock \bibinfo{journal}{\emph{arXiv preprint arXiv:2102.04925}}
  (\bibinfo{year}{2021}).
\newblock


\bibitem[Wu et~al\mbox{.}(2022)]%
        {wu2022linkteller}
\bibfield{author}{\bibinfo{person}{Fan Wu}, \bibinfo{person}{Yunhui Long},
  \bibinfo{person}{Ce Zhang}, {and} \bibinfo{person}{Bo Li}.}
  \bibinfo{year}{2022}\natexlab{}.
\newblock \showarticletitle{Linkteller: Recovering private edges from graph
  neural networks via influence analysis}. In \bibinfo{booktitle}{\emph{2022
  IEEE Symposium on Security and Privacy (SP)}}. IEEE,
  \bibinfo{pages}{2005--2024}.
\newblock


\bibitem[Xiao et~al\mbox{.}(2020)]%
        {xiao2020adversarial}
\bibfield{author}{\bibinfo{person}{Taihong Xiao}, \bibinfo{person}{Yi-Hsuan
  Tsai}, \bibinfo{person}{Kihyuk Sohn}, \bibinfo{person}{Manmohan Chandraker},
  {and} \bibinfo{person}{Ming-Hsuan Yang}.} \bibinfo{year}{2020}\natexlab{}.
\newblock \showarticletitle{Adversarial learning of privacy-preserving and
  task-oriented representations}. In \bibinfo{booktitle}{\emph{Proceedings of
  the AAAI Conference on Artificial Intelligence}}, Vol.~\bibinfo{volume}{34}.
  \bibinfo{pages}{12434--12441}.
\newblock


\bibitem[Xu et~al\mbox{.}(2018)]%
        {xu2018dpne}
\bibfield{author}{\bibinfo{person}{Depeng Xu}, \bibinfo{person}{Shuhan Yuan},
  \bibinfo{person}{Xintao Wu}, {and} \bibinfo{person}{HaiNhat Phan}.}
  \bibinfo{year}{2018}\natexlab{}.
\newblock \showarticletitle{DPNE: Differentially private network embedding}. In
  \bibinfo{booktitle}{\emph{PAKDD}}. Springer, \bibinfo{pages}{235--246}.
\newblock


\bibitem[Yang et~al\mbox{.}(2021)]%
        {yang2021financial}
\bibfield{author}{\bibinfo{person}{Shuo Yang}, \bibinfo{person}{Zhiqiang
  Zhang}, \bibinfo{person}{Jun Zhou}, \bibinfo{person}{Yang Wang},
  \bibinfo{person}{Wang Sun}, \bibinfo{person}{Xingyu Zhong},
  \bibinfo{person}{Yanming Fang}, \bibinfo{person}{Quan Yu}, {and}
  \bibinfo{person}{Yuan Qi}.} \bibinfo{year}{2021}\natexlab{}.
\newblock \showarticletitle{Financial risk analysis for SMEs with graph-based
  supply chain mining}. In \bibinfo{booktitle}{\emph{Proceedings of the
  Twenty-Ninth International Conference on International Joint Conferences on
  Artificial Intelligence}}. \bibinfo{pages}{4661--4667}.
\newblock


\bibitem[Zhang et~al\mbox{.}(2022)]%
        {zhang2022inference}
\bibfield{author}{\bibinfo{person}{Zhikun Zhang}, \bibinfo{person}{Min Chen},
  \bibinfo{person}{Michael Backes}, \bibinfo{person}{Yun Shen}, {and}
  \bibinfo{person}{Yang Zhang}.} \bibinfo{year}{2022}\natexlab{}.
\newblock \showarticletitle{Inference attacks against graph neural networks}.
  In \bibinfo{booktitle}{\emph{Proceedings of the 31th USENIX Security
  Symposium}}. \bibinfo{pages}{1--18}.
\newblock


\bibitem[Zhang et~al\mbox{.}(2021)]%
        {zhang2021graphmi}
\bibfield{author}{\bibinfo{person}{Zaixi Zhang}, \bibinfo{person}{Qi Liu},
  \bibinfo{person}{Zhenya Huang}, \bibinfo{person}{Hao Wang},
  \bibinfo{person}{Chengqiang Lu}, \bibinfo{person}{Chuanren Liu}, {and}
  \bibinfo{person}{Enhong Chen}.} \bibinfo{year}{2021}\natexlab{}.
\newblock \showarticletitle{GraphMI: Extracting Private Graph Data from Graph
  Neural Networks}.
\newblock \bibinfo{journal}{\emph{arXiv preprint arXiv:2106.02820}}
  (\bibinfo{year}{2021}).
\newblock


\bibitem[Zhou et~al\mbox{.}(2020)]%
        {zhou2020privacy}
\bibfield{author}{\bibinfo{person}{Jun Zhou}, \bibinfo{person}{Chaochao Chen},
  \bibinfo{person}{Longfei Zheng}, \bibinfo{person}{Xiaolin Zheng},
  \bibinfo{person}{Bingzhe Wu}, \bibinfo{person}{Ziqi Liu}, {and}
  \bibinfo{person}{Li Wang}.} \bibinfo{year}{2020}\natexlab{}.
\newblock \showarticletitle{Privacy-Preserving Graph Neural Network for Node
  Classification}.
\newblock \bibinfo{journal}{\emph{arXiv preprint arXiv:2005.11903}}
  (\bibinfo{year}{2020}).
\newblock


\bibitem[Zhu et~al\mbox{.}(2020)]%
        {zhu2020beyond}
\bibfield{author}{\bibinfo{person}{Jiong Zhu}, \bibinfo{person}{Yujun Yan},
  \bibinfo{person}{Lingxiao Zhao}, \bibinfo{person}{Mark Heimann},
  \bibinfo{person}{Leman Akoglu}, {and} \bibinfo{person}{Danai Koutra}.}
  \bibinfo{year}{2020}\natexlab{}.
\newblock \showarticletitle{Beyond homophily in graph neural networks: Current
  limitations and effective designs}.
\newblock \bibinfo{journal}{\emph{Advances in Neural Information Processing
  Systems}}  \bibinfo{volume}{33} (\bibinfo{year}{2020}),
  \bibinfo{pages}{7793--7804}.
\newblock


\end{thebibliography}

\end{document}